\documentclass[11pt,a4paper,oneside,titlepage=true]{amsart}
\usepackage{geometry}
\usepackage{cite}
\usepackage{amsmath,amssymb,amsfonts,amsthm}
\usepackage{algorithm}
\usepackage{algpseudocode}
\usepackage{graphicx}
\usepackage{textcomp}
\usepackage{caption}
\usepackage{subcaption}
\usepackage{tikz}
\usepackage{float}
\usetikzlibrary{arrows}

\newcommand{\cN}{\mathcal{N}}
\newcommand{\cA}{\mathcal{A}}
\newcommand{\cR}{\mathcal{R}}
\newcommand{\RR}{\mathbb{R}}
\usepackage{mathtools}
\usepackage{url}
\newtheorem{definition}{Definition}

\tikzset{
  treenode/.style = {align=center, inner sep=0pt, text centered,
    font=\sffamily},
  arn_n/.style = {treenode, circle, white, font=\sffamily\bfseries, draw=black,
    fill=black, text width=1.5em},
  arn_r/.style = {treenode, circle, red, draw=red, 
    text width=1.5em, very thick},
  arn_x/.style = {treenode, rectangle, draw=black,
    minimum width=0.5em, minimum height=0.5em}
}

\title[Feature Importance and Trends in Random Forests]{A Notion of Feature Importance by Decorrelation and Detection of Trends by Random Forest Regression}

\author{Yannick Gerstorfer, Lena Krieg, Max Hahn-Klimroth}
\address{Frankfurt Institute for Advanced Studies, Goethe University Frankfurt, Frankfurt 60325 Germany (e-mail: gerstorfer@fias.uni-frankfurt.de)}
\address{Faculty of Computer Sciences, TU Dortmund University, Dortmund 44227 Germay (e-mail: lena.krieg@tu-dortmund.de}
\address{Faculty of Computer Sciences, TU Dortmund University, Dortmund 44227 Germay (e-mail: maximilian.hahnklimroth@tu-dortmund.de}
\thanks{All authors were supported by the German Research Council, grant DFG FOR 2975.}
\begin{document}

\begin{abstract}
In many studies, we want to determine the influence of certain features on a dependent variable.
More specifically, we are interested in the strength of the influence -- i.e., is the feature relevant? -- and, if so, how the feature influences the dependent variable.
Recently, data-driven approaches such as \emph{random forest regression} have found their way into applications (Boulesteix et al., 2012). 
These models allow to directly derive measures of feature importance, which are a natural indicator of the strength of the influence.
For the relevant features, the correlation or rank correlation between the feature and the dependent variable has typically been used to determine the nature of the influence.
More recent methods, some of which can also measure interactions between features, are based on a modeling approach.
In particular, when machine learning models are used, SHAP scores are a recent and prominent method to determine these trends (Lundberg et al., 2017).

In this paper, we introduce a novel notion of feature importance based on the well-studied Gram-Schmidt decorrelation method. 
Furthermore, we propose two estimators for identifying trends in the data using random forest regression, the so-called absolute and relative transversal rate.
We empirically compare the properties of our estimators with those of well-established estimators on a variety of synthetic and real-world datasets.
\end{abstract}

\maketitle

\section{Introduction}
\label{sec:introduction}
In many studies, scientific researchers are faced with high-dimensional but limited data to determine the influence of specific features on a dependent variable. 
Typically, the data consist of both numerical and categorical features, and strong artificial multivariate correlations appear. 
In particular, when data are generated from observations of live animals or collected in medical procedures, it is very likely that the data are unbalanced and, even worse, not all combinations of features contain samples. Therefore, it is unlikely that all necessary assumptions of classical statistical tests will be met. 
Machine learning methods have gained popularity among researchers because they can produce robust effect estimates with minimal assumptions.
A plain but prominent example is the \emph{random forest regression}.
Due to advances in data science concepts as well as the increasing computational power available to any research group, such data-driven approaches are finding their way into life science studies \cite{Boulesteix_2012}. 
Random forest regression, as all machine learning models, makes few assumptions about the distributions of the underlying data and is particularly robust to noise and outliers. 
Finally, it allows to directly derive measures of feature importance, which are a natural indicator of the strength of influence of individual features \cite{Fraser_1965,Beraha_2019,Parveen_2012}. 
In cases where classical statistical tools such as {ANOVA} can be applied, it is well known that most features found to be significant by ANOVA also have high feature importance and vice versa \cite{Chicco_2021,Saarela_2021}.

Once relevant features have been found, it is important to determine how the values of the features affect the dependent variable.
Probably the oldest approach is to measure the correlation or rank correlation between a feature and the dependent variable.
More recent methods, some of which can also measure interactions between features, are based on a modeling approach.
A model (e.g. a multivariate linear regression model) is trained and its parameters can be used to determine trends, especially when machine learning models are used, the SHAP scores \cite{Shapley53} are a recent and prominent method to determine these trends.
These approaches use the model rather than the raw data.
This can help to identify trends that are not directly visible in the data, but are hidden behind noise.
On the other hand, a decent model is required so that these trends are reliable.

The goal of this paper is twofold.
First, since dependencies between features are known to influence feature importance scores, we introduce a notion of feature importance based on the well-studied Gram-Schmidt decorrelation method.
This notion is empirically compared with a similar approach based on residual learning and the classical impurity-based feature importance and permutation importance.
Second, we propose two estimators to identify trends in the data using random forest regression.
We exploit the structure of random forests, i.e. at each split node we can compare the average prediction in the left and right subtrees.
Since the left subtree is built on data below a threshold and the right subtree contains data above that threshold, this induces a natural estimator of some kind of correlation between the feature and the predicted variable.

\section{Background and Notation}
\subsection{Feature Importance}
With respect to random forests, two types of feature importance scores are well known in the literature. 
The first one is an \emph{impurity-based} feature importance. 
The so-called impurity is quantified by the splitting criterion of the collection of contained decision trees. 
Therefore, it is likely to overestimate the importance of large numerical features (if the dataset is not standardized). 
Furthermore, it is possible that features that may not be predictive on unseen data are found to be important in the case of overfitting.
For these reasons, a second type of feature importance, the so-called \emph{permutation importance}, has found its way into the literature and is to be preferred \cite{Breiman_2001}.
It is defined as the decrease in model performance when a single feature is randomly shuffled.
Of course, this permutation-based approach has its shortcomings - in particular, if there are clusters of (highly) correlated \cite{Breiman_2001} features.
One approach to overcome this problem, which is often used in the process of feature extraction, is to keep only one variable per cluster \cite{Chen_2020, Louppe_2014,Guyon_2002}. 
If the ultimate goal is to design a decent prediction model with as few features as possible, this is the state of the art. 
But in some cases, researchers are actually more interested in estimating the importance of each feature to determine which features influence the dependent variable and how strongly. 
In this setting, it may be convenient to treat the correlations differently. 
There are at least two \emph{decorrelation techniques} that are usually used either for clustering data or for designing well-performing prediction models: the Gram-Schmidt decorrelation technique \cite{Kun_Zhang_2006_GramSchmidt} or residual-based decorrelation \cite{Dezfouli_2019}.
The main idea in both cases is to subtract the information from a given feature $F_i$ given by $F_1, \ldots, F_{i-1}, F_{i+1},\ldots F_d$ and use this \emph{residual} to train the model.

\subsection{Trends}
We compare three different ways to define \emph{trends} in the data set.
The simplest way one might think of examining a trend between the values of a feature $Y$ and the predicted variable is to use the \emph{correlation coefficient} 
$r(X,Y) := \mathrm{Cov}(X,Y) / (\sigma(X) \sigma(Y))$, which reflects linear trends. A more general correlation coefficient that handles any \emph{monotone} trends are various types of rank correlation coefficients such as the \emph{Spearman correlation coefficient} $\rho(X,Y) = r(R(X), R(Y))$ where $R(\cdot)$ denotes the rank function.
This method of finding trends is well established and only considers the observable raw data.

Another approach does not look at the raw data, but fits a model and looks for trends in that model.
Many practitioners tend to identify trends in multivariate tasks by fitting a linear model to the data and interpreting the sign and corresponding $p-$value of the coefficient of a feature as a trend. 
We will denote this coefficient by $r_{LM}(F)$. 
However, we will see that this can be very misleading, even for very simple data sets.

In recent years, an old concept from mathematical game theory, called \emph{Shapley values}, has been used to interpret machine learning models \cite{Lundberg2017}.
In particular, they are well understood mathematically for tree-based models and random forests.
The Shapley value of a feature with respect to a data point measures how much the feature value contributes to the prediction compared to the average prediction, and is defined as the average marginal contribution of the feature value among all possible combinations of features. 
For a formal definition, see Shapley's original paper \cite{Shapley53}, and for a detailed discussion of how to use the concept in machine learning, see \cite{pmlr-v108-janzing20a, pmlr-v119-sundararajan20b}. 
Clearly, these Shapley scores can be used to determine trends.

\subsection{Studied Datasets}
To test the performance of our estimators in practice, we use two very well-known real datasets, called \emph{Kaggle fish market dataset} (FISH) \cite{kaggle_fish} and \emph{California housing data} (HOUSING) \cite{kaggle_housing}.
In addition, we create three different synthetic datasets to explore certain aspects of the estimators.

FISH contains the records of seven different common fish species in fish market sales. 
The features are species, weight, vertical length, diagonal length, transverse length, height, and width for each fish. 
Of these characteristics, we used weight, height, and width to predict vertical length.
The California housing data refers to the houses found in a given California county and summary statistics based on 1990 census data. 
The features are longitude, latitude, median age of the house, total number of rooms, total number of bedrooms, population, number of households, median income, and ocean proximity for each county with median house value as the prediction target.
We transformed the ocean proximity feature into an ordinal scale.

The first synthetic data set (SYN1) is derived from a base data set $B$ consisting of 1000 samples and 10 features, 3 of which are informative. 
The base dataset is standardised by removing the mean and scaling to unit variance. It is then combined with a noise dataset $N$ standardised the same way and with the same structure but no informative features. 
A family of data sets is obtained by $$D_w = (1 - w) B + w N \quad (w \in [0.01, 0.02, \dots, 1.]). $$ 
SYN1 consists of the combination of the base data set with 250 different random noise data sets. 
SYN1 is used to compare the robustness of trend estimators.

The second synthetic data set (SYN2) consists of 100 samples with independently generated features 
\begin{align*}
    X_0 \sim 3 \cdot \cN(0,1), \qquad X_1 \sim 2 \cdot \cN(0,1), \qquad X_2 \sim \cN(0,1).
\end{align*}
Furthermore, given $X_0$, we define 
\begin{align*}
    A_0 = X_0 + \cN(0,1), \qquad A_1 = X_0 + 10 \cdot \cN(0,1), \qquad A_2 = X_0^2 + 10 \cdot \cN(0,1).
\end{align*}
The true label is given by $$Y = 4 X_0^{1.5} + 2 X_1 + 0.5 X_2^2.$$ 
Thus, the real labels depend on $X_0, X_1, X_2$ and $A_0, A_1, A_2$ can be considered as noisy instances of $X_0$ with different types of dependencies. SYN2 is used to compare different notions of feature importance.

The third synthetic data set (SYN3) consists of 100 samples with only one informative feature $X_0$, defined as previously.
Moreover, $A_0, A_1, A_2$ are defined as above.
The true labels are now given by $Y = 4 X_0^{1.5}.$ 
SYN3 is used to compare the notions of feature importance on a \emph{cluster} of correlated features, in direct comparison to SYN2, in which two additional, uncorrelated, informative features are present.

\section{Contribution}
\subsection{Finding Trends in a Dataset}
We compare the commonly used regression coefficients $r, \rho$, the linear model-based trend estimator, a Shapley-based trend estimator, and propose two novel estimators based on random forest regression to determine the trends of features.
For this purpose, we simply define the Shapley-based trend of a feature as the correlation between its values $X$ and its Shapley values $s(X)$, so that we obtain the estimators $r(X, s(X))$ and $\rho(X, s(X))$, respectively. 

The two proposed trend estimators are the absolute and the relative \emph{traversal rate}. 
The random forest regression model uses an ensemble of uncorrelated decision trees. 
At each node, the current data set is partitioned into two partition classes based on the values of the node's feature. 
We assume without loss of generality that the data in the \emph{left} partition class belong to small feature values and the data in the \emph{right} partition class belong to large feature values. 
To determine the trend of a feature $F$, it seems reasonable to compare the mean of the features in the left and right partition classes per node. 
If the average value of the predicted variable in the left tree is smaller than in the right tree, this corresponds to a positive correlation with the feature $F$.
More formally, let $\{ F_j \}_{j = 1 \ldots n}$ denote the set of nodes in the random forest in which the data is partitioned with respect to feature $F$.
The corresponding partition classes are called $L(F_n)$ and $R(F_n)$. If the feature $F$ is clear from the context, we abbreviate these classes to $L_n$ and $R_n$.
Furthermore, for a subset $A$ of the values of the predicted variable, we define $AVG(A) = \lvert{A}\rvert^{-1} \underset{a\in A}{\sum} a$ as the average value of the set $A$.
This allows us to define our trend estimators.
\begin{definition}
Given a random forest $\mathcal{R}$, let $\{ F_j \}_{j = 1 \ldots n}$ denote the set of nodes in the random forest in which the data is split with respect to feature $F$. 
The absolute transversal rate of feature $F$ is defined as
\begin{align*}
    ATR(\mathcal{R}, F) = n^{-1} \sum_{i = 1}^n \left( \mathbf{1} \{ AVG( L(F_i) ) < AVG( R(F_i) ) \} - \mathbf{1} \{ AVG( L(F_i) ) > AVG( R(F_i) ) \}\right). 
\end{align*}
Moreover, the relative transversal rate of feature $F$ is defined as
\begin{align*}
    RTR(\mathcal{R}, F) = 2 \sum_{i = 1}^n \frac{ R(F_i) - L(F_i)}{\lvert L(F_i) + R(F_i) \rvert}. 
\end{align*}
\end{definition}
The ATR formalizes the idea that we have a trend when the feature with a higher value causes the model to return a higher value.
The RTR also takes into account the relative difference between the average values in the partition classes.

\begin{figure}[H]
    \centering
    \begin{tikzpicture}[->,>=stealth',level/.style={sibling distance = 5cm/#1,
      level distance = 1.5cm}]
    \node [arn_r] {$F_1$}
        child{ node [arn_n] {} 
                child{ node [arn_n] {} 
                    child{ node [arn_n] {2} edge from parent node[above left] {}
                    }
    				child{ node [arn_n] {4} }
            }
            child{ node [arn_r] {$F_2$}
    				child{ node [arn_n] {7} }
    				child{ node [arn_n] {3} }
            }                            
        }
        child{ node [arn_r] {$F_3$}
                child{ node [arn_n] {} 
    				child{ node [arn_n] {8}  }
    				child{ node [arn_n] {12} }
                }
                child{ node [arn_n] {}
    				child{ node [arn_n] {4} }
    				child{ node [arn_n] {6} }
                }
    		}
    ;
    \end{tikzpicture}
    \caption{Each occurrence of feature $F$ splits the dataset into two parts. In the example, $F_1$ creates partition classes $L_1 = \{ 2, 4, 7, 3 \}$ and $R_1 = \{ 8, 12, 4, 6 \}$. The split at $F_2$ creates classes $L_2 = \{ 7 \}$ and $R_2 = \{ 3\}$, whereas the split at $F_3$ defines $L_3 = \{ 8, 12 \}$ and $R_3 = \{ 4, 6 \}$.}\label{fig:traversal}
\end{figure}

\subsubsection*{Trend Estimator}
For the empirical analysis of the datasets, we wrote a \emph{trend estimator module}, which trains both a random forest regression model as well as a linear model on the input data. For each feature, the trend estimator then outputs 
\begin{enumerate}
    \item the coefficient of the linear model
    \item Spearman's rank correlation and Pearson correlation coefficient between the target and
    \begin{enumerate}
        \item the Shapley value
        \item the feature
    \end{enumerate}
    \item the absolute traversal rate (ATR)
    \item the relative traversal rate (RTR).
\end{enumerate} 

\subsection{Measures of Importance}
We compare four different notions of \emph{feature importance}.
Two definitions, the impurity-based feature importance and the permutation-based feature importance, are well studied objects \cite{Breiman_2001}.
We use \emph{scikit-learn's} default implementation of these measures.

In addition, we introduce two novel types of feature importance based on \emph{residual learning}.
The idea is that the importance of feature $F_i$ is determined by its residuals given features $F_1, \ldots, F_{i-1}, F_{i+1},\ldots F_d$.
With a slight misuse of notation, we interpret $F_i \in \RR^n$ as the vector of all values corresponding to feature $F_i$ and denote by $Y \in \RR^n$ the values of the dependent variable.
We denote by $\cA_j$ an arbitrary algorithm that takes $F_1, \ldots, F_{j-1}$ as input and outputs a vector in $\RR^n$.
Given a fixed permutation $\pi$ of $[d]$, we denote by $i_1, \ldots, i_d$ the new order under $\pi$. 
To determine the importance of $F_i$, we determine its importance under all permutations $\pi$ with the property that $i_d = i$ and weight it by the performance of a model consisting only of the feature $F_i$.
The interpretation is as follows: given all other features, what can be learned from feature $F_{i_d}$?
The algorithm to compute the importance can now be expressed as follows.
\begin{itemize}
    \item For all permutations $\pi$ which map $i \mapsto d$, do the following
    \begin{itemize}
        \item Define $W^\pi_1 = F_{i_1}$.
        \item Replace the values of feature $F_{i_j}$ with $W^{\pi}_{i_j} = F_{i_j} - \mathcal{A_{i_j}}(W^{\pi}_{i_1}, \ldots, W^\pi_{i_{j-1}})$ (for $j = 2 \ldots d$).
        \item Train a random forest with features $\{ W_{i_{j}} \}$.
        \item Determine the impurity-based feature importance of $W^{\pi}_{i_d}$.
    \end{itemize}    
    \item Determine the average feature importance of $F_i$ as the mean over all $W^{\pi}_{i_d}$, call this $\mathrm(FI)_i$.
    \item Train a random forest regressor $\cR_i$ with feature $F_i$ and dependent variable $Y$ and measure $r(\cR(F_i), Y)$.
    \item Return $\tilde f_i = r(\cR_i(F_i), Y) \mathrm(FI)_i$
\end{itemize}
After applying this algorithm, we are left with $\tilde f_1, \ldots, \tilde f_d$. 
Finally, we define the feature importance based on the residual algorithm $\cA$ as the standardized version of the above estimator, namely
$$ f_j(\cA) = \frac{\tilde f_j}{\sum_{i=1}^d \tilde f_i}.$$
Formally, the algorithm is given as Algorithm \ref{alg:cap}.
We note the following.
\begin{itemize}
    \item $f_j(\cA)$ is a random quantity because it depends on the training of the random forest regressors $\cR_1, \ldots, \cR_d$ and the random forest regressors using the features $\{ W_{i_{j}} \}$.
    \item In applications, it may not be possible to iterate over all permutations $\pi$. Instead, the average impurity-based feature importance is estimated by sampling some permutations.
    \item The algorithm is highly dependent on the residual algorithm $\cA$.
\end{itemize}

\begin{algorithm}
\caption{Residual-based feature importance}\label{alg:cap}
{\small
\begin{algorithmic}
\Require $d$ features $F_1, \ldots, F_d,$ residual algorithm $\cA$, dependent variable $Y$
\State $S_d \gets$ set of permutations of $\{ 1, 2, \ldots, d\}$
\State $\mathrm{FeatImp} \gets (0, ..., 0) \in \mathbb{R}^d$
\For{ $\pi \in S_d$ }
    \State $F_{i_{j}} \in \mathbb{R}^{n} \gets$ $j-$th feature vector under permutation $\pi$
    \State $W^\pi_{i_1} = F_{i_1}$ 
    \For{ $j = 2 \ldots d$ } 
        \State $W^\pi_{i_j} \gets F_{i_j} - \mathcal{A}_{i_j}(W^{\pi}_{i_1}, \ldots, W^\pi_{i_{j-1}})$ 
    \EndFor
    \State $\mathrm{RF} \gets$ generate a random forest model with features $W^\pi_{i_1}, \ldots, W^\pi_{i_d}$ and dependent variable $Y$
    \State $\mathrm{FI} \gets $ result of impurity-based feature importance of $\mathrm{RF}$ for feature $W^{\pi}_{i_d}$
    \State $\mathrm{\cR_{i_d}} \gets $ generate random forest model with feature $F_{i_d}$ and dependent variable $Y$
    \State $\mathrm{FeatImp}[k] \gets \mathrm{FeatImp}[k] + r(\cR_{i_d}(F_{i_d}), Y) \cdot \mathrm{FI}$ 
\EndFor
\State $\mathrm{FeatImp} \gets \frac{\mathrm{FeatImp}}{ || \mathrm{FeatImp} ||_1}$
\State \Return $\mathrm{FeatImp}$
\end{algorithmic}
}
\end{algorithm}

In this contribution, we empirically analyze the feature importance based on two different residual algorithms: classical residual learning by random forest regression and decorrelation by the Gram-Schmidt method.

\subsubsection*{Residual Learning-based Feature Importance}
Following \cite{Dezfouli_2019}, it is a natural idea to define the family of residual algorithms $\cA_2, \ldots, \cA_{d}$ as a family of random forest regressors.
More precisely, given $W^{\pi}_{i_1}, \ldots, W^\pi_{i_{j-1}}$, we train a random forest regressor $\cR$ on those features with the dependent variable $F_{i_j}$.
Hence,
$$ \cA_j(W^{\pi}_{1}, \ldots, W^\pi_{j-1}) = \cR( W^\pi_1, \ldots, W^\pi_{j-1}).$$
Thus, we subtract from $F_{i_j}$ everything that can be learned by random forest regressors from the first $j-1$ features under $\pi$.
This approach is classically known as residual learning and finds prominent applications in machine learning \cite{resnet}.

\subsubsection*{Gram-Schmidt decorrelation-based Feature Importance}
Another natural approach is to use the very famous Gram-Schmidt orthogonalization technique. 
While it has been used in mathematics for a good century to generate orthogonal bases of vector spaces, it was first applied in the early 2000s to find independent components in complex data sets \cite{Kun_Zhang_2006_GramSchmidt}.
The most important observation is that the covariance is an inner product, so the very general Gram-Schmidt orthogonalization technique can be applied with the covariance to create decorrelated features.
Here, we define
$$ \cA_j(W^{\pi}_{1}, \ldots, W^\pi_{j-1}) = \sum_{i = 1}^{j-1} \frac{\mathrm{Cov}(F_j, W^\pi_i)}{\mathrm{Cov}(F_j, F_j)} F_j.$$
A major advantage may be that this orthogonalization method, unlike the above approach, is fully mathematically understandable.
However, it may be brittle to nonlinear dependencies.

\section{Results}
\subsection{Finding Trends}
In the following, we report our empirical results on the performance of the different trend estimators on the HOUSING, FISH, and SYN1 datasets.
\subsubsection{SYN1}
\begin{figure}[ht]
    \centering
    \begin{tabular}{ccccc}
         \includegraphics[width=0.19\textwidth]{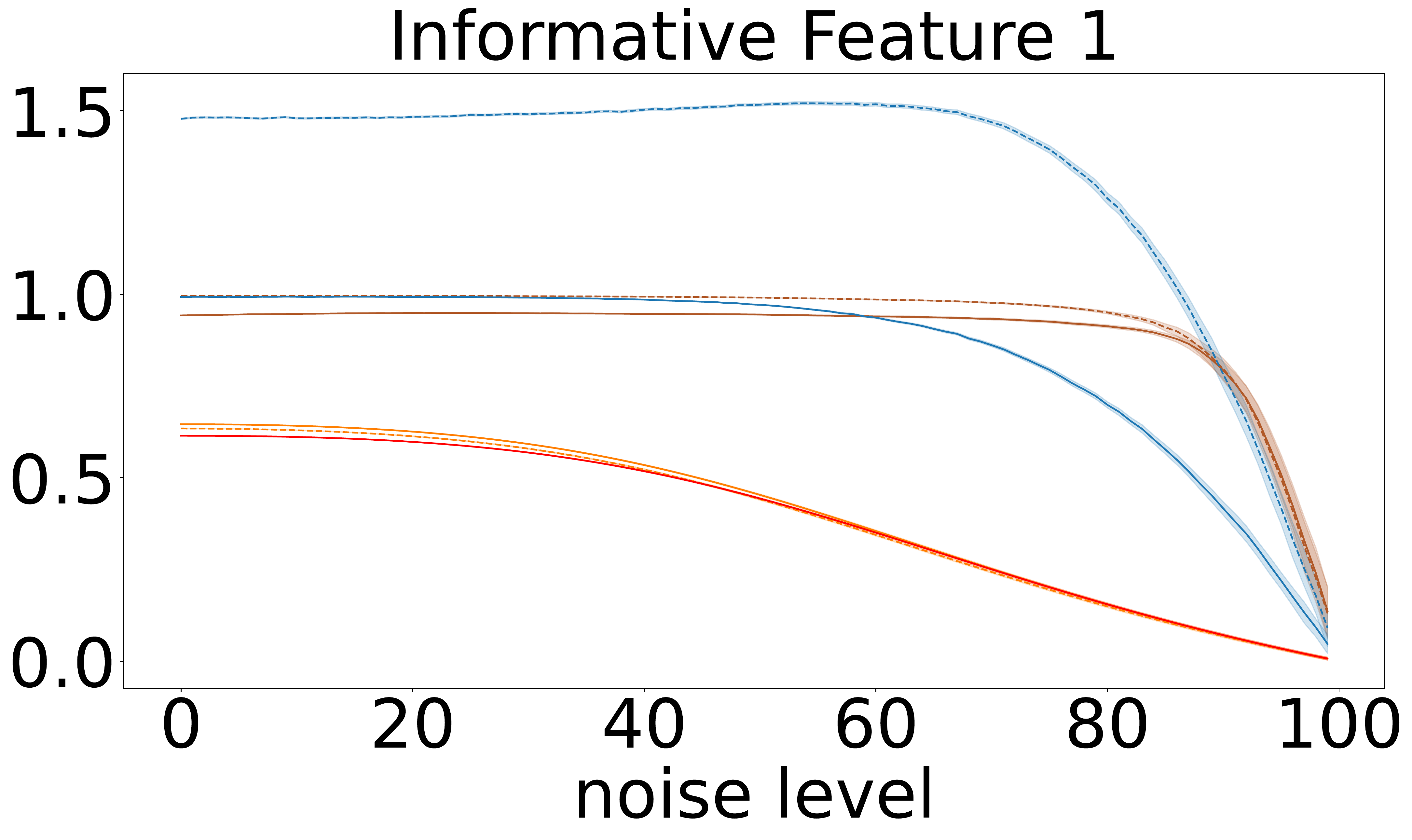} & \includegraphics[width=0.19\textwidth]{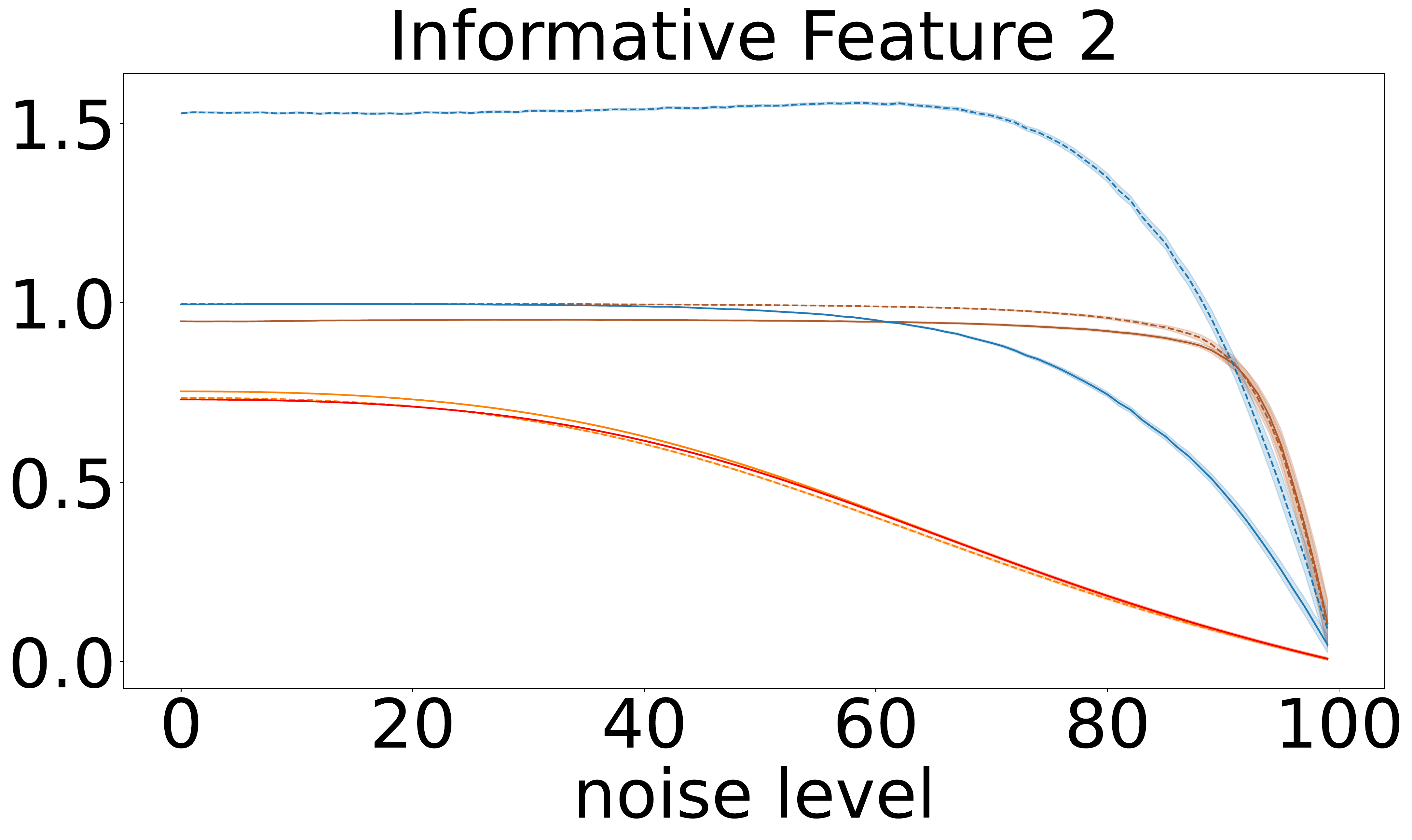} & \includegraphics[width=0.19\textwidth]{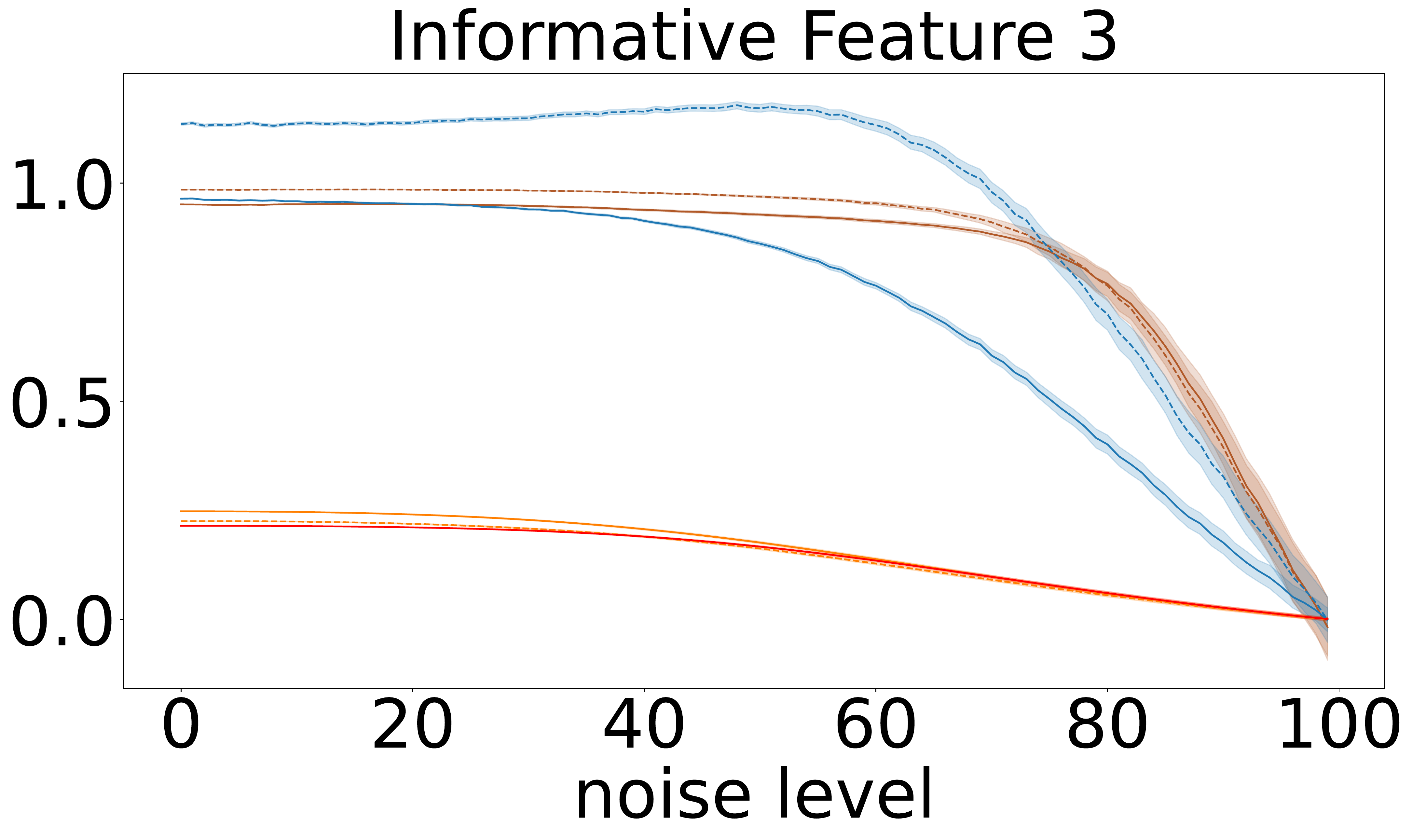} & \includegraphics[width=0.19\textwidth]{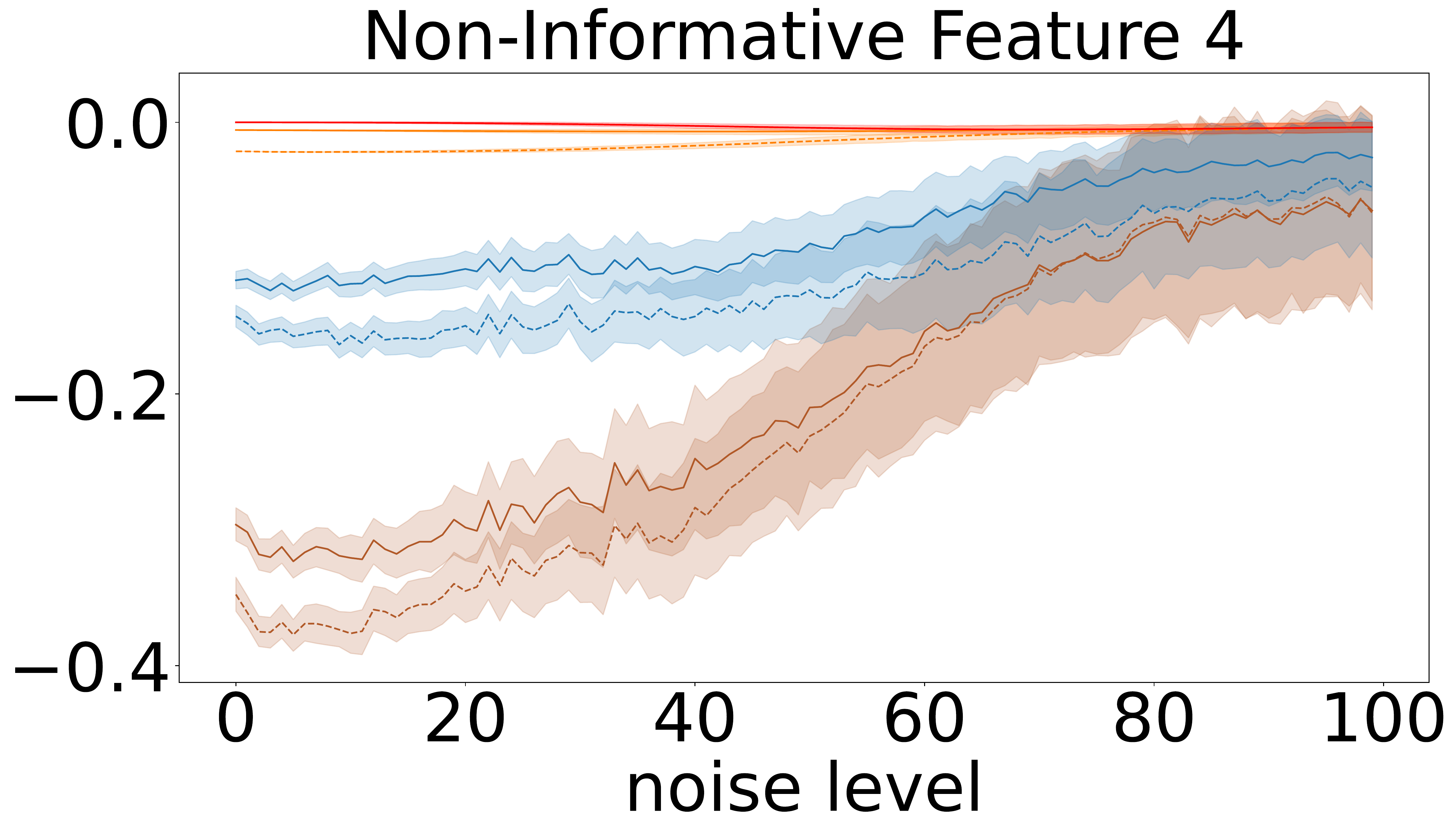} & \includegraphics[width=0.19\textwidth]{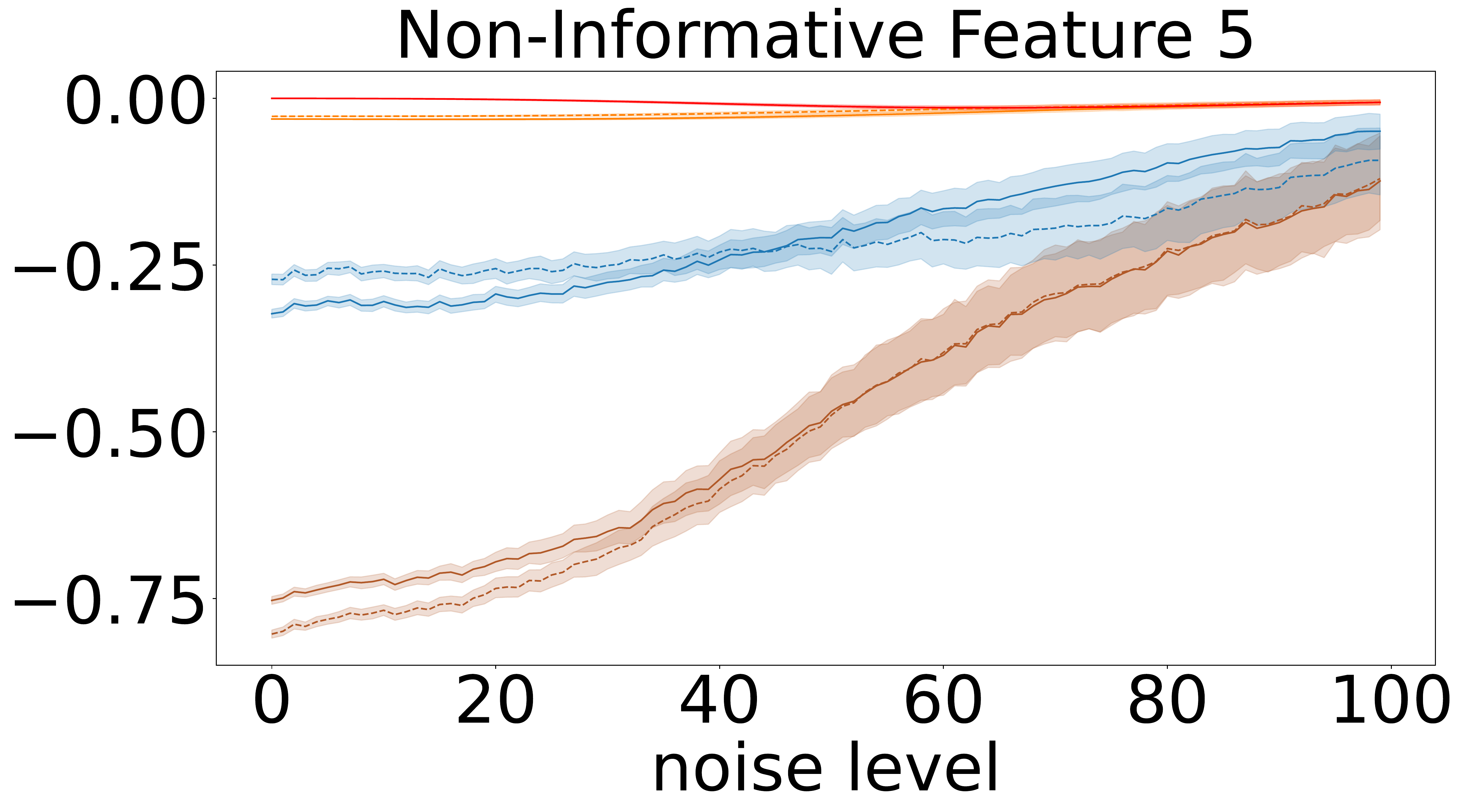} 
         \tabularnewline
         \includegraphics[width=0.19\textwidth]{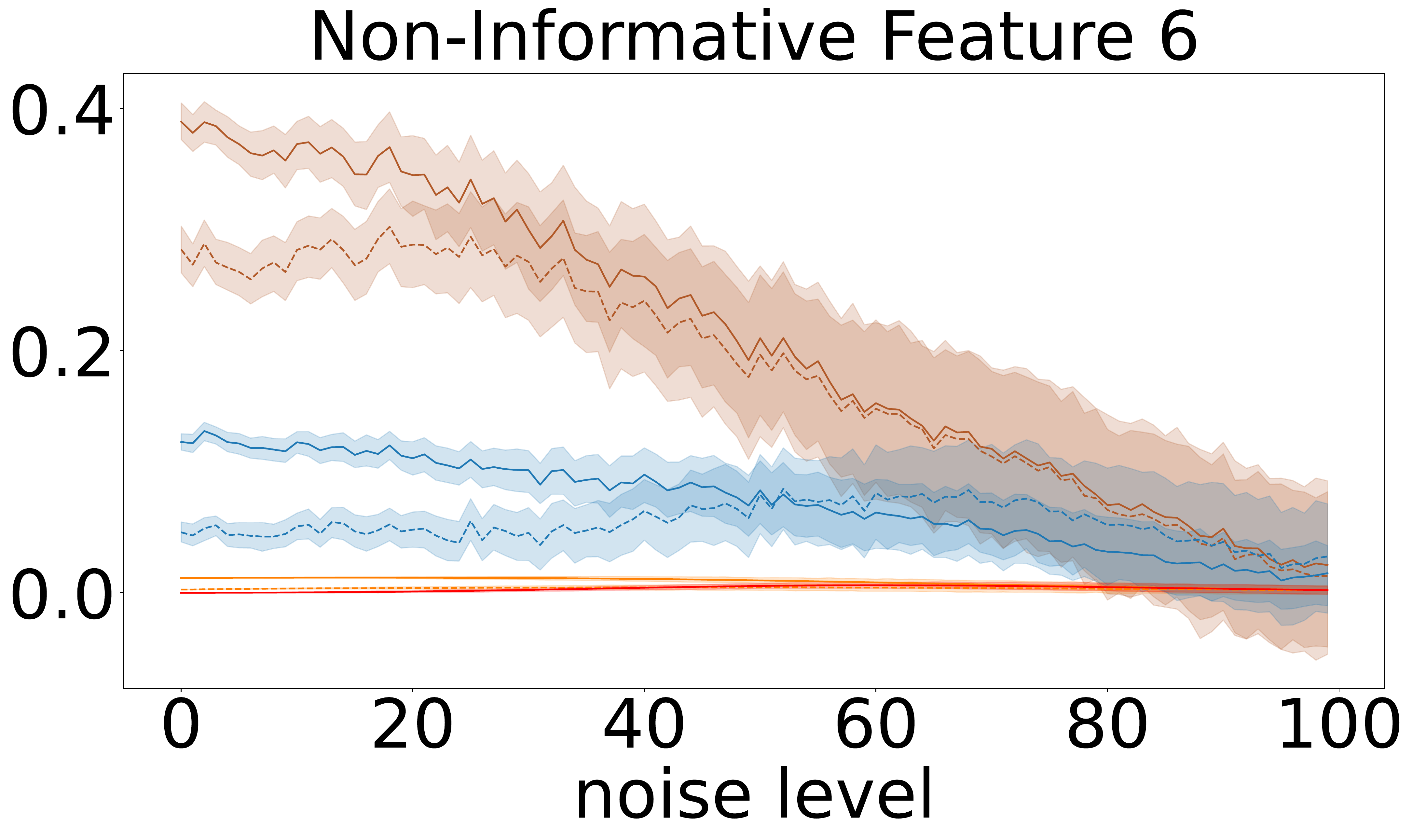} & \includegraphics[width=0.19\textwidth]{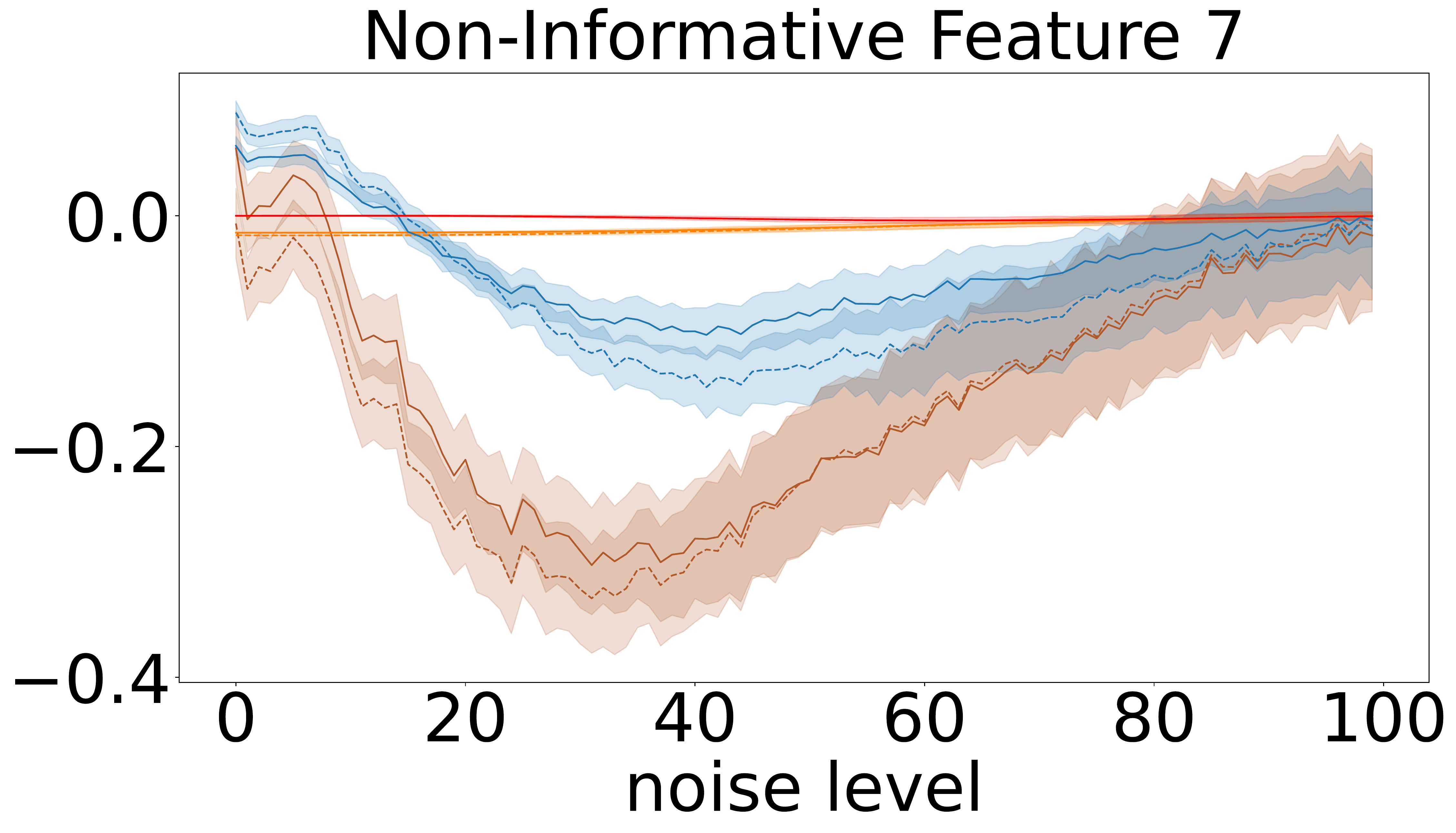} & \includegraphics[width=0.19\textwidth]{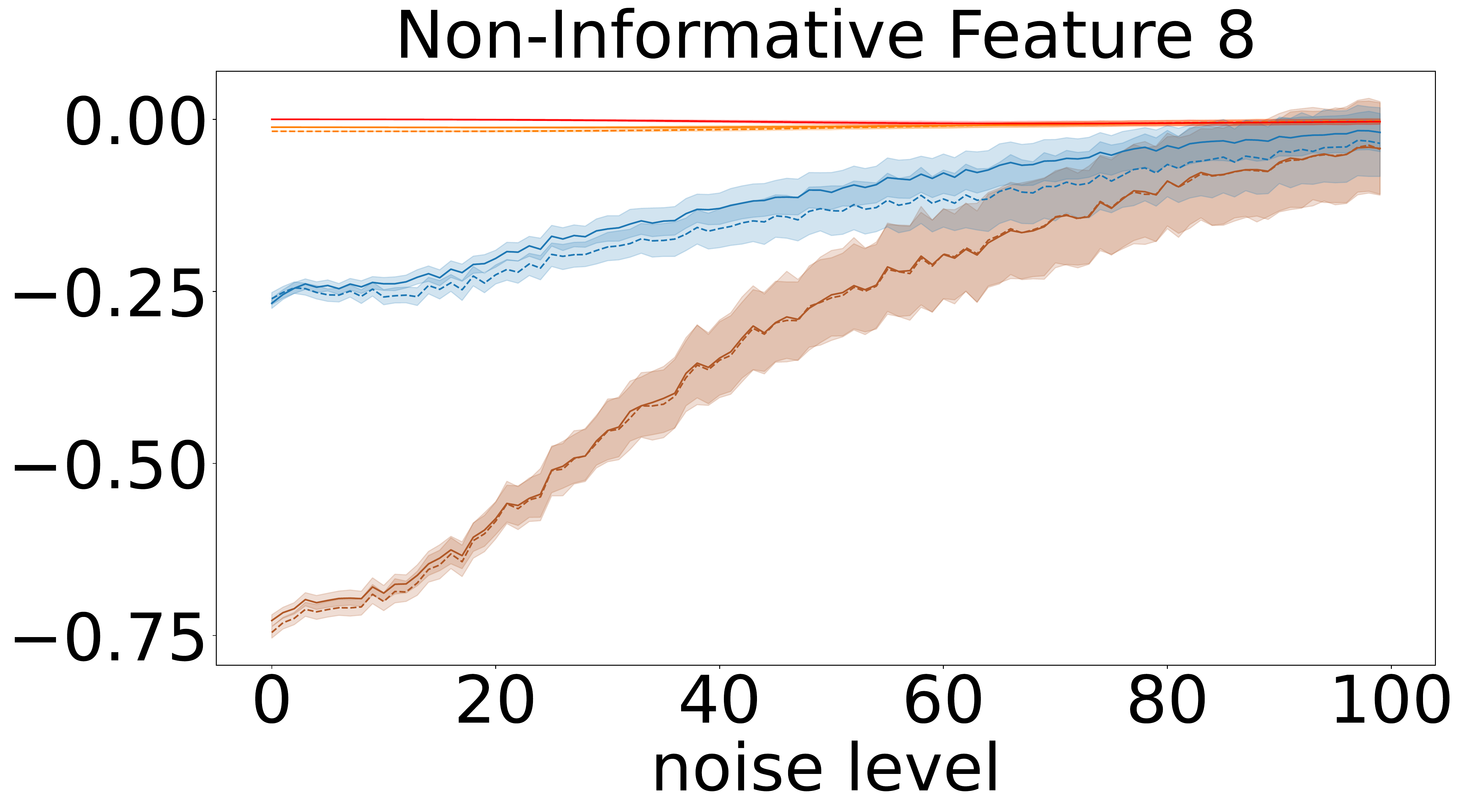} & \includegraphics[width=0.19\textwidth]{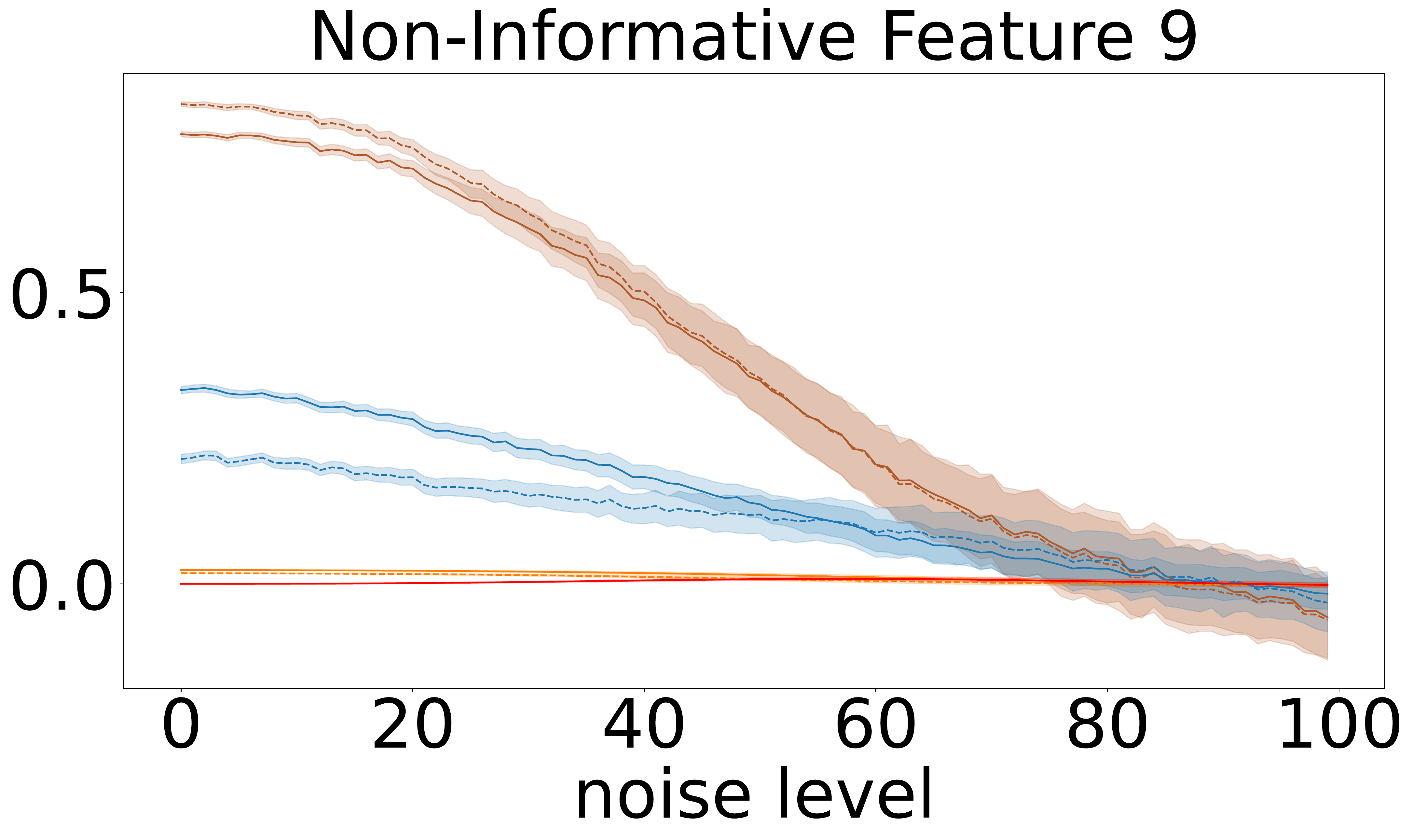} & \includegraphics[width=0.19\textwidth]{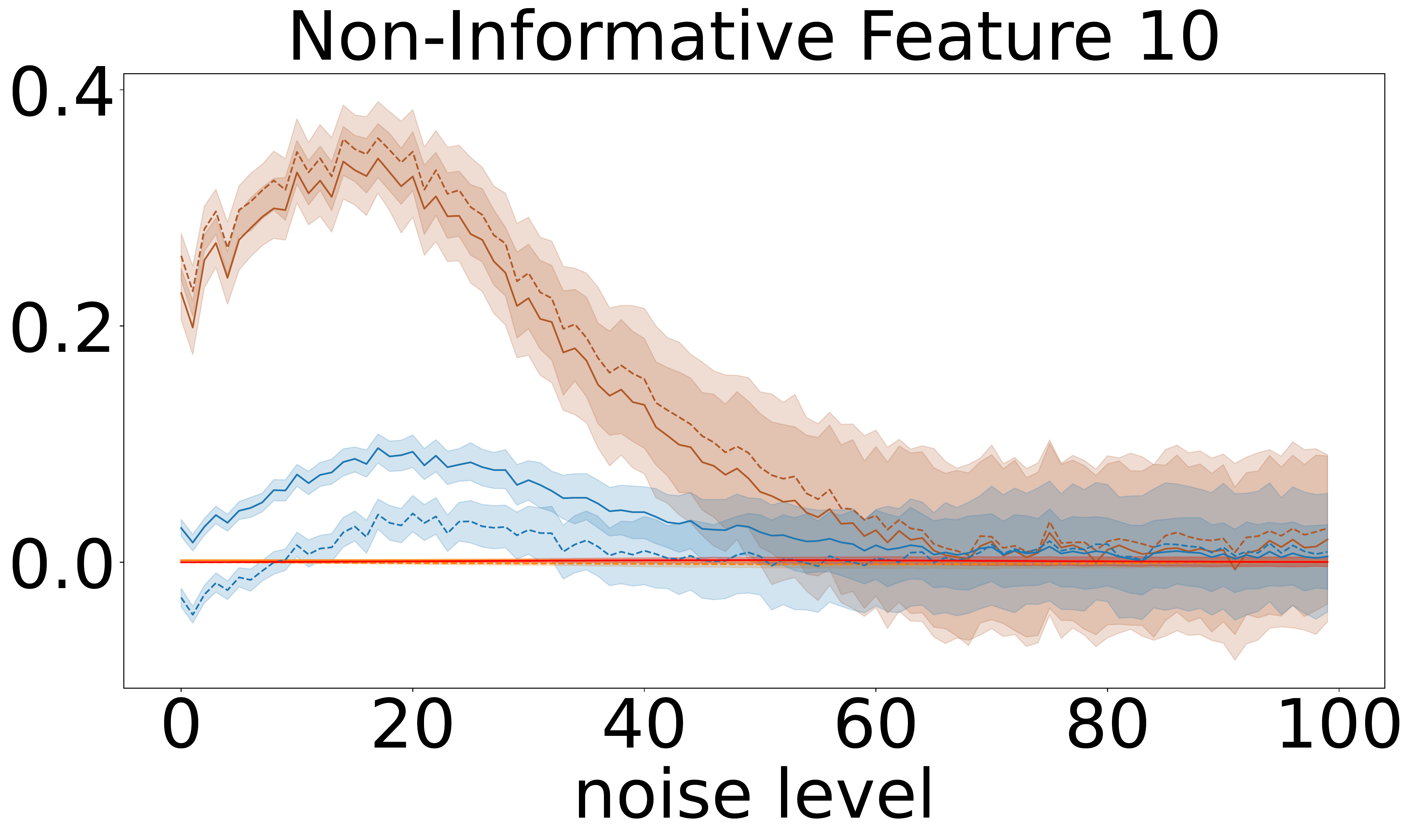} 
         \tabularnewline
         \multicolumn{5}{c}{\includegraphics[width=0.5\textwidth]{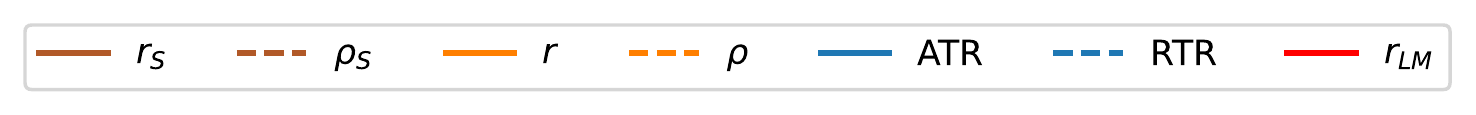}}\tabularnewline
    \end{tabular}
    \caption{Mean and 95\% confidence interval for the different trend estimators on SYN1 for 250 independent trials. On the $x-$axis, the proportion of noise is reported. Features 1-3 are informative, whereas features 4-10 are non-informative.}\label{fig:noise}
\end{figure}

This dataset was used to test the robustness of the different trend estimators with respect to the mixing of the dataset with noise.
To do this, the trend estimator module was applied to $D_w$ for each $w \in [0.01, \ldots, 1.]$ for each of the 250 random noise data sets. In our experiment, the aggregated output shows that both ATR and RTR, as well as the Shapley correlation, are more robust than the linear model for the informative features (Fig. \ref{fig:noise}). The Shapley values are the most robust, followed by RTR and ATR.
Interestingly, non-informative features were also assigned large $\rho_S, r_S$, RTR and ATR values.

\subsubsection{FISH}
\begin{figure}[ht]
    \centering
    \includegraphics[width=0.7\textwidth]{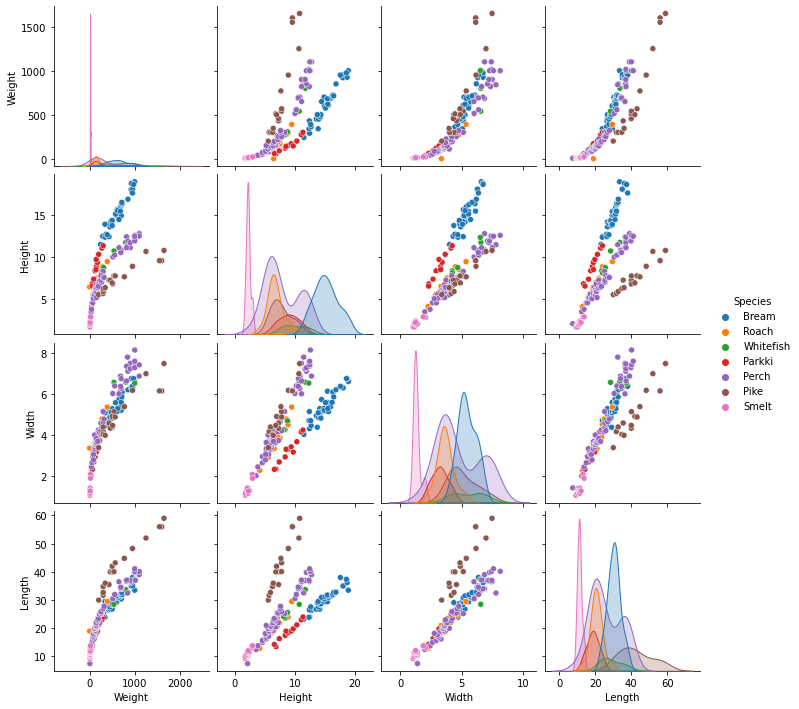}
    \caption{Pairplot of the used fish market dataset features (Weight, Height and Width) and the predicted variable (Length).}\label{fig:pairplot_fish}
\end{figure}

We performed three experiments on the FISH dataset using the features \emph{Weight}, \emph{Height} and \emph{Width} to predict \emph{Length}. All three selected features are positively correlated with the target (Fig. \ref{fig:pairplot_fish}).

\begin{figure}[ht]
    \centering
        \begin{tabular}{ccc}
            \includegraphics[width=0.32\textwidth]{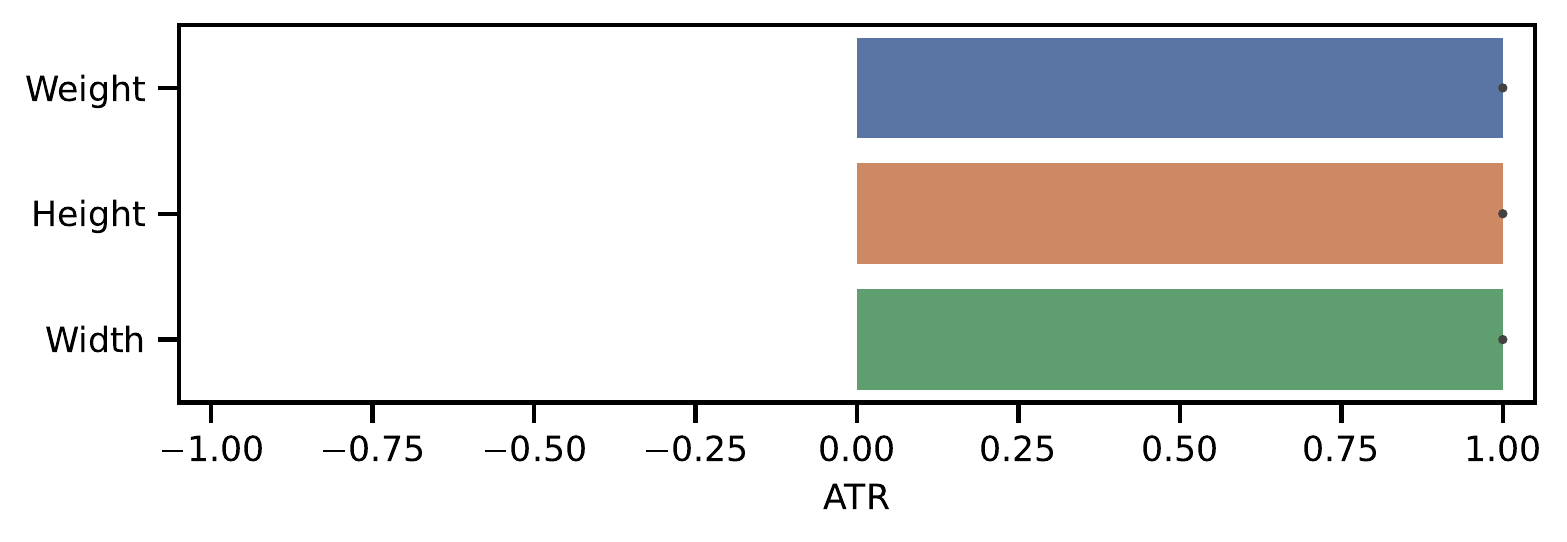} & \includegraphics[width=0.32\textwidth]{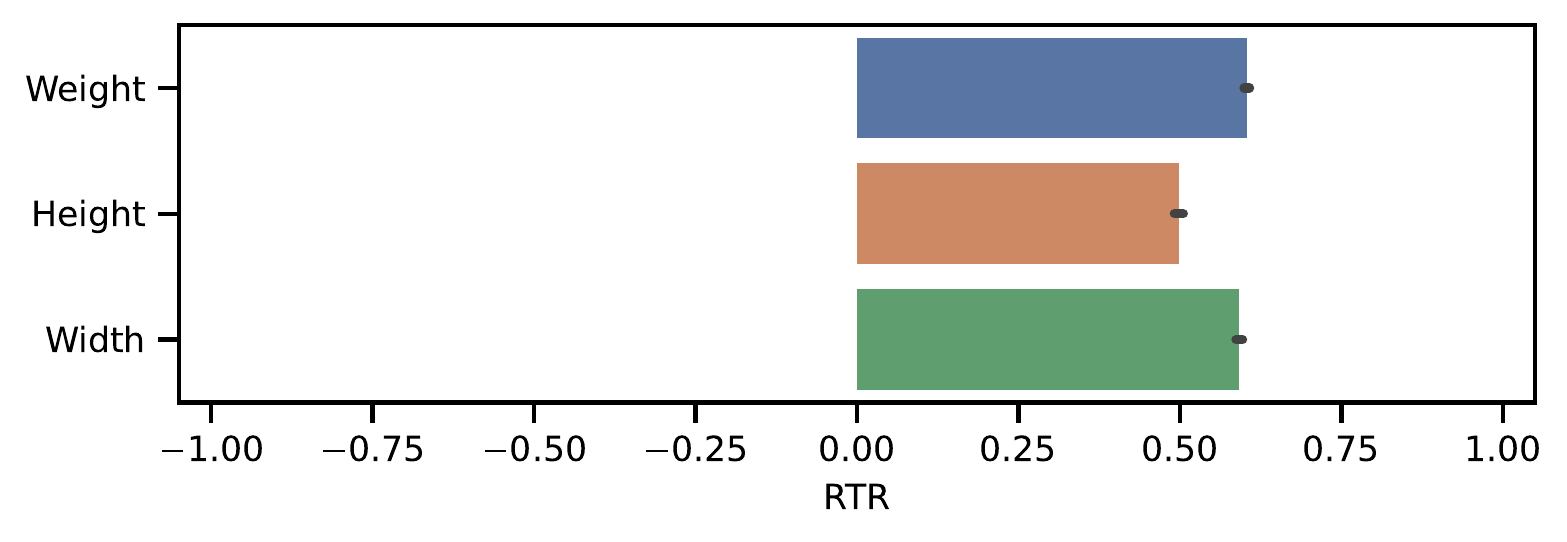} & \includegraphics[width=0.32\textwidth]{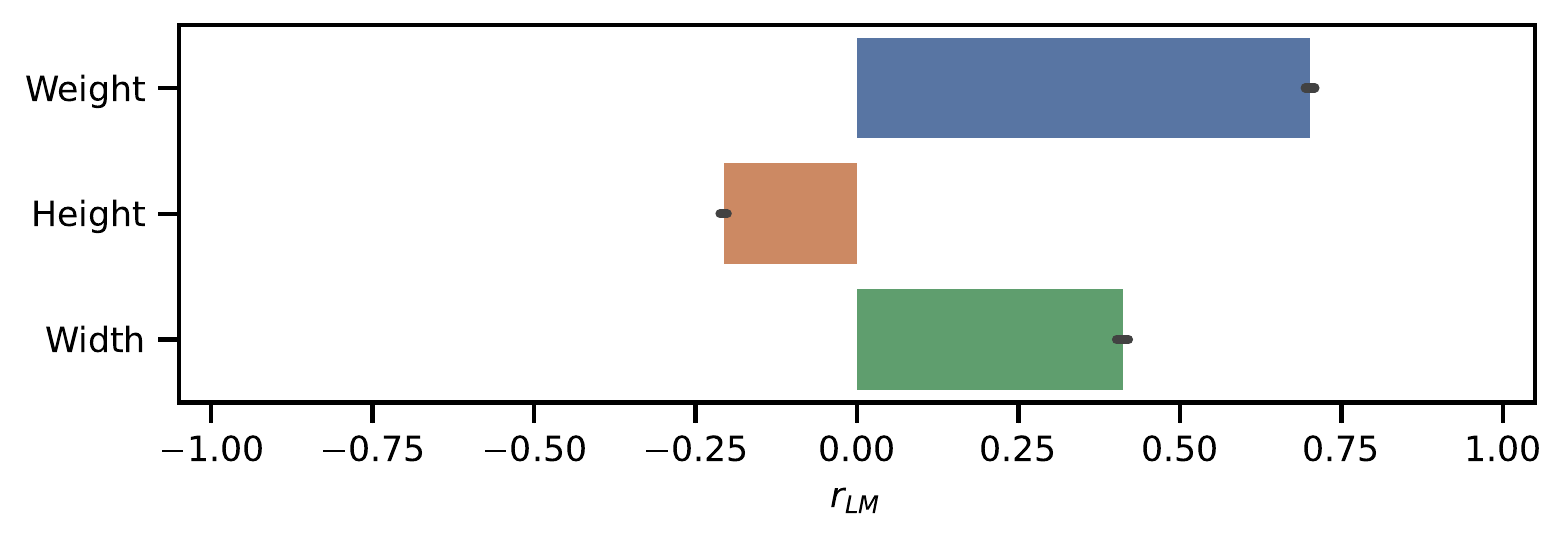}
            \tabularnewline
             \includegraphics[width=0.32\textwidth]{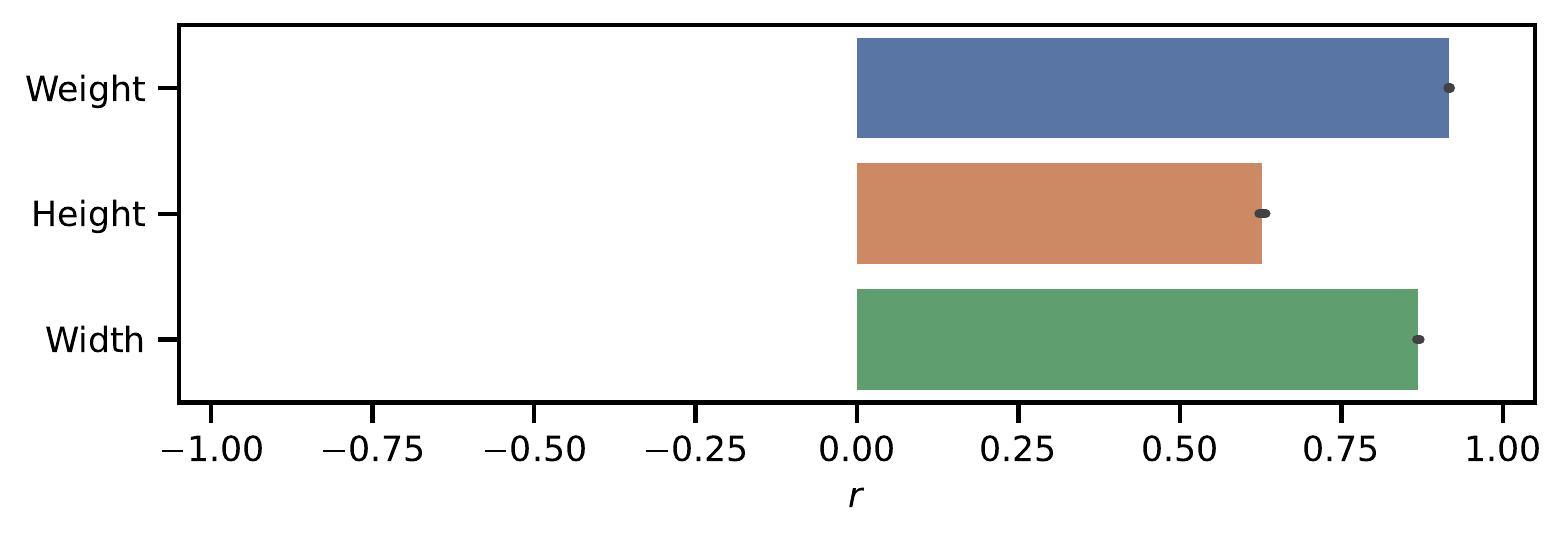}  & \includegraphics[width=0.32\textwidth]{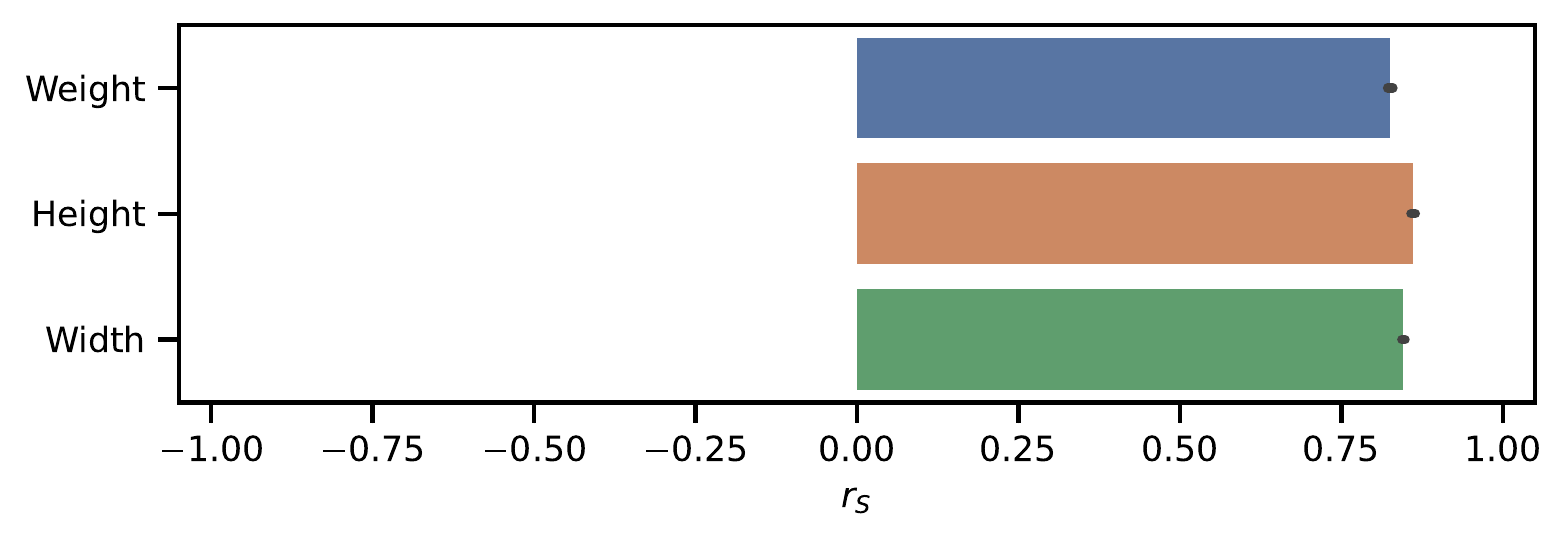} & \includegraphics[width=0.32\textwidth]{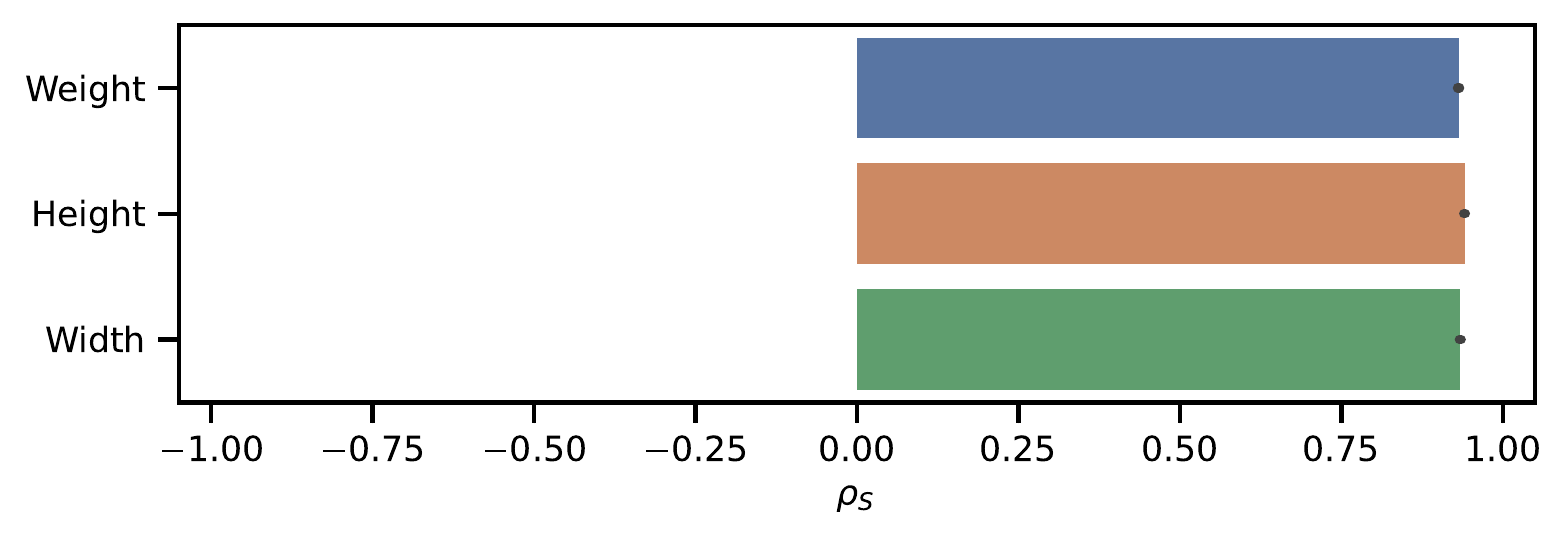}
             \tabularnewline
        \end{tabular}
        
    \caption{Comparison of the trend estimators for FISH. We report the mean and the standard deviation of the different trend estimators over 100 bootstrap iterations, each containing 70\% of the data.}
    \label{fig:barplot_fish_2}
\end{figure}

First, we applied the trend estimation module to the FISH dataset. To control for random effects, we performed 100 bootstrapping iterations, sampling from a subset of 70 \%. The linear regression model assigned a negative coefficient to the \emph{Height} feature, while the other trend estimators reported a positive trend (Figure \ref{fig:barplot_fish_2}).

\begin{figure}[ht]
    \centering    
    \includegraphics[width=0.32\textwidth]{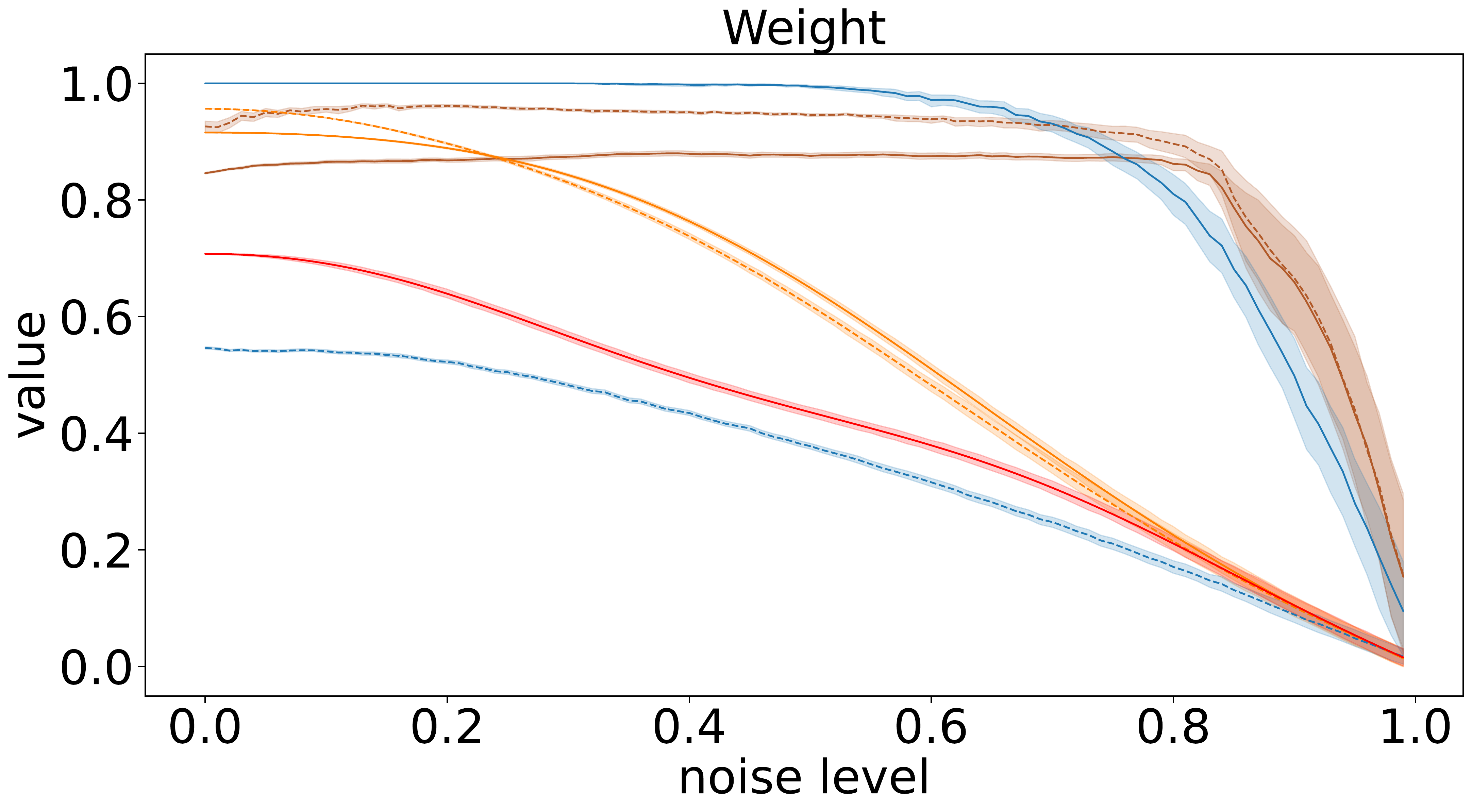}
    \includegraphics[width=0.32\textwidth]{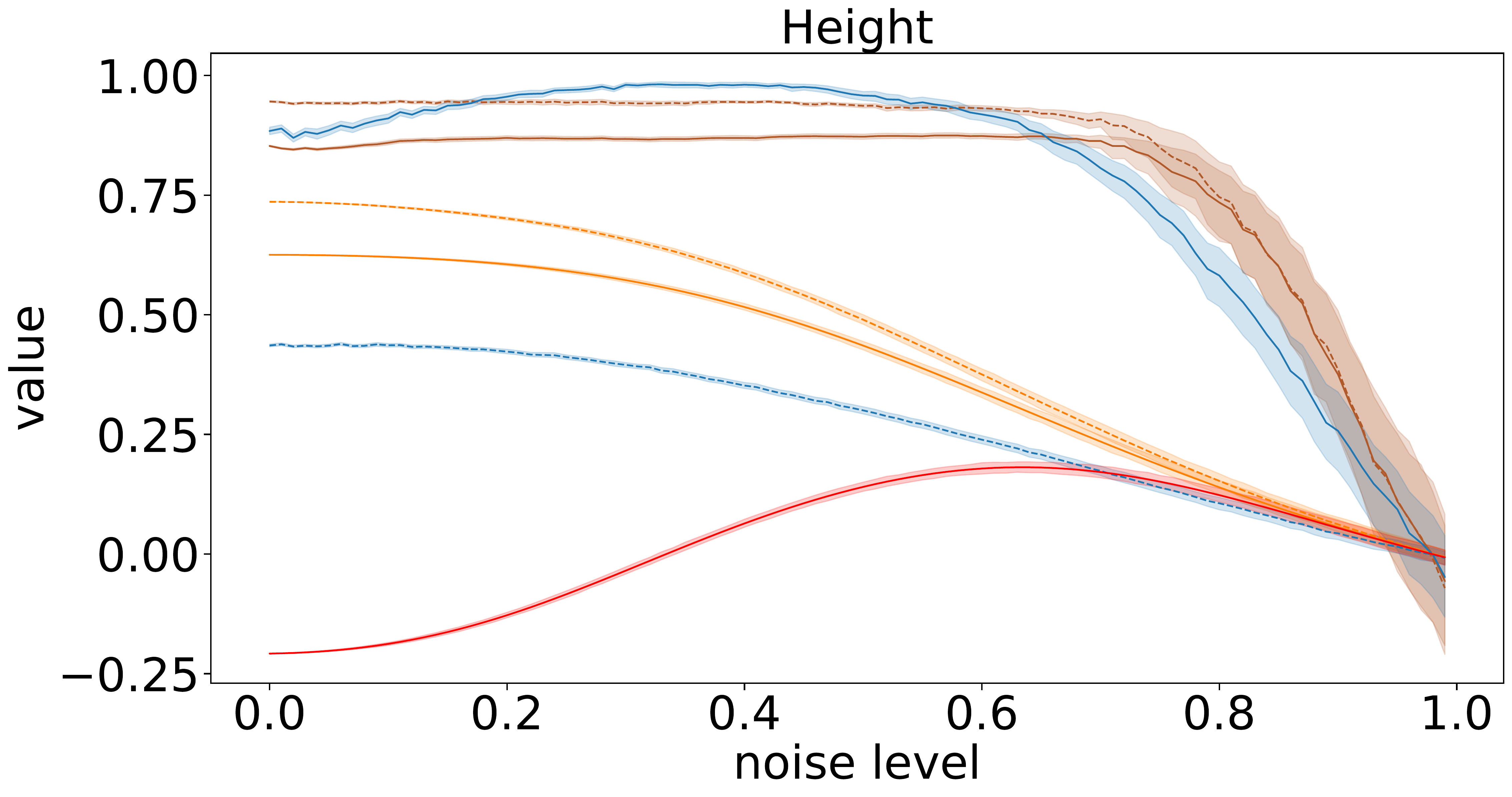}
    \includegraphics[width=0.32\textwidth]{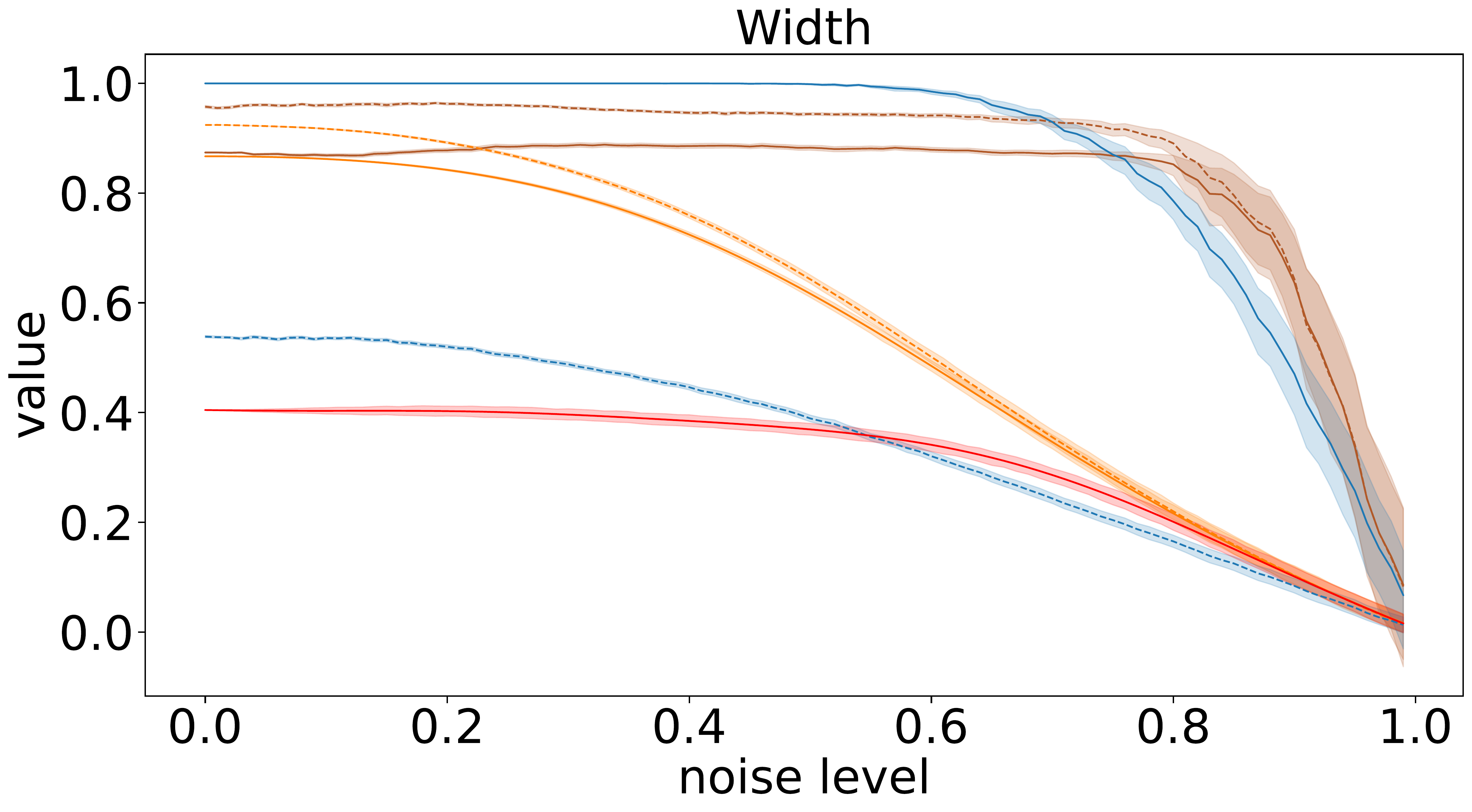}
    \includegraphics[width=0.5\textwidth]{images/noise/legend.pdf}
    \caption{Mean and 95\% confidence interval w.r.t. 100 independent iterations over noise on FISH. The $x-$axis reports the proportion of noise mixed to the real data.}\label{fig:fish_noise}
\end{figure}

To evaluate the robustness of the trend estimators to noise, we used a random mixing strategy similar to that used to create SYN1. The FISH data were standardized and mixed with random noise ranging from $0\%$ to $99\%$ noise before being used as input to the trend estimator module. We found that the linear model and the RTR became unstable as the feature-to-target correlation $r$ and $\rho$ decreased, while the ATR and the Shapley measures $r_S$ and $\rho_S$ remained relatively unaffected up to much higher mixing rates (Fig. \ref{fig:fish_noise}).

\subsubsection{HOUSING}
\begin{figure}[ht]
    \centering
    \begin{tabular}{ccc}
        \includegraphics[width=0.49\textwidth]{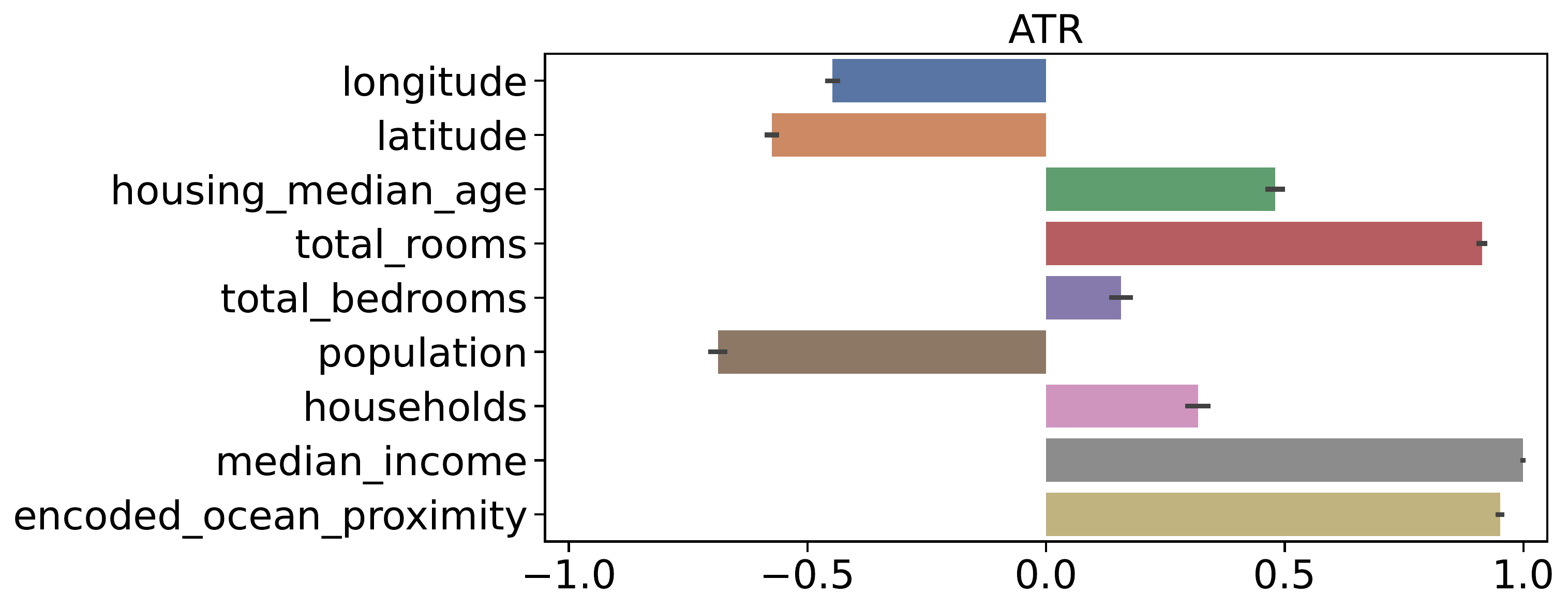} & \includegraphics[width=0.49\textwidth]{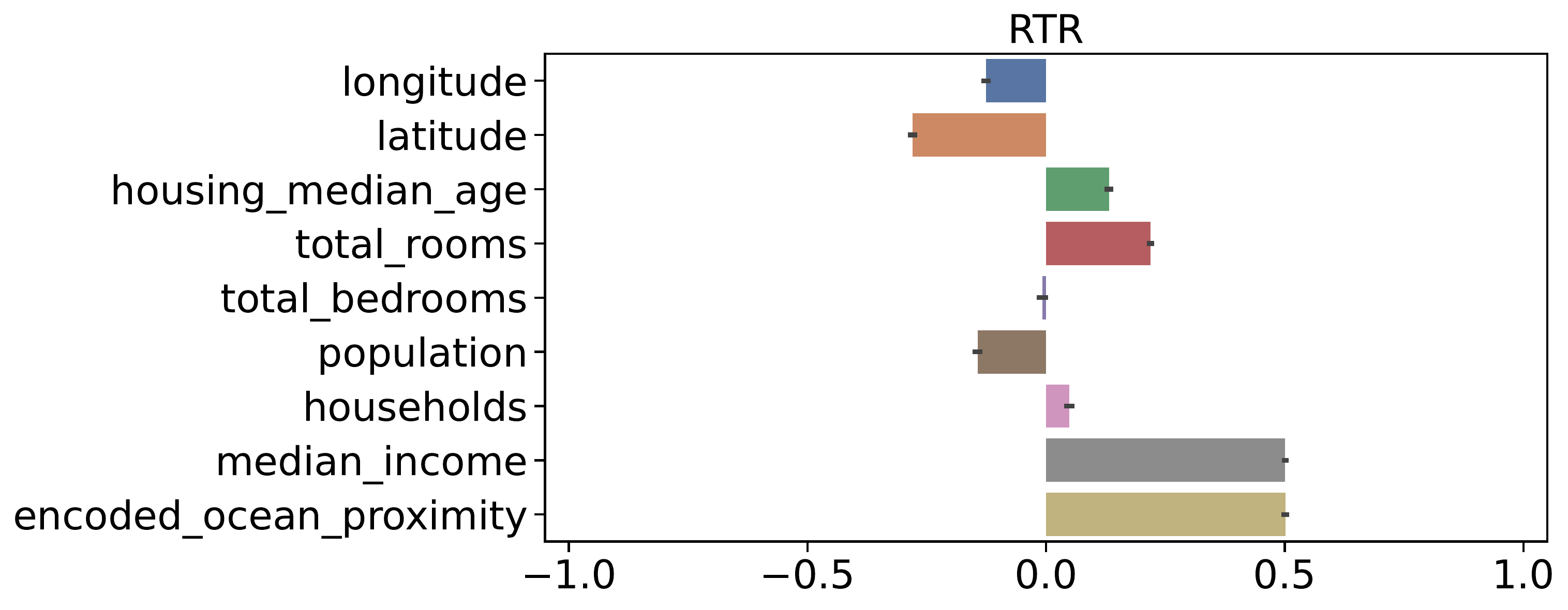}
        \tabularnewline
        \includegraphics[width=0.49\textwidth]{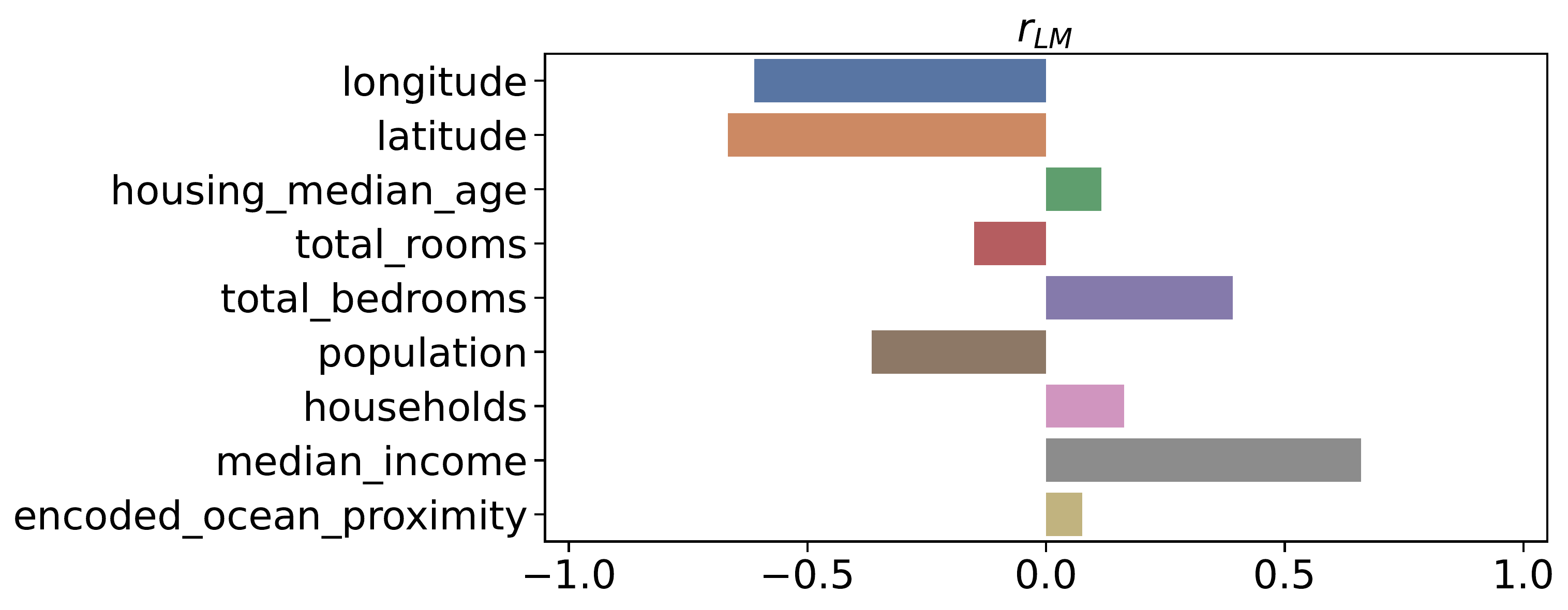} & \includegraphics[width=0.49\textwidth]{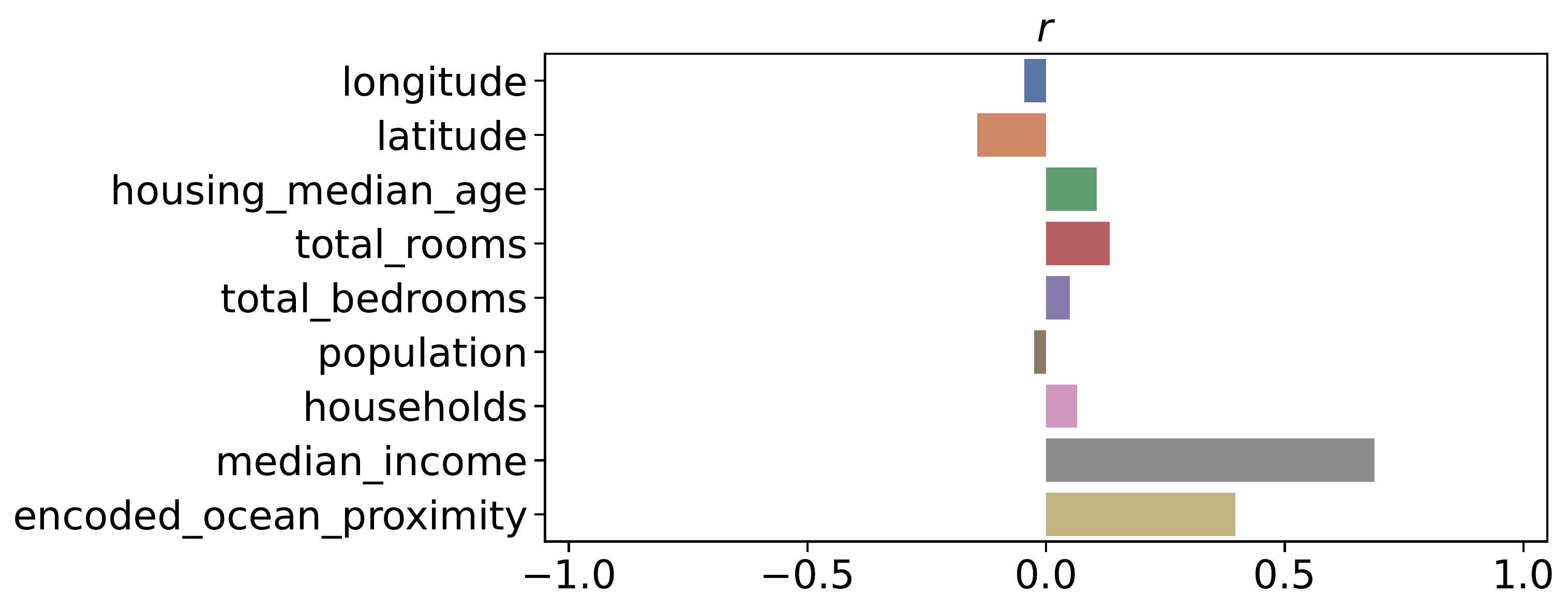}
        \tabularnewline
        \includegraphics[width=0.49\textwidth]{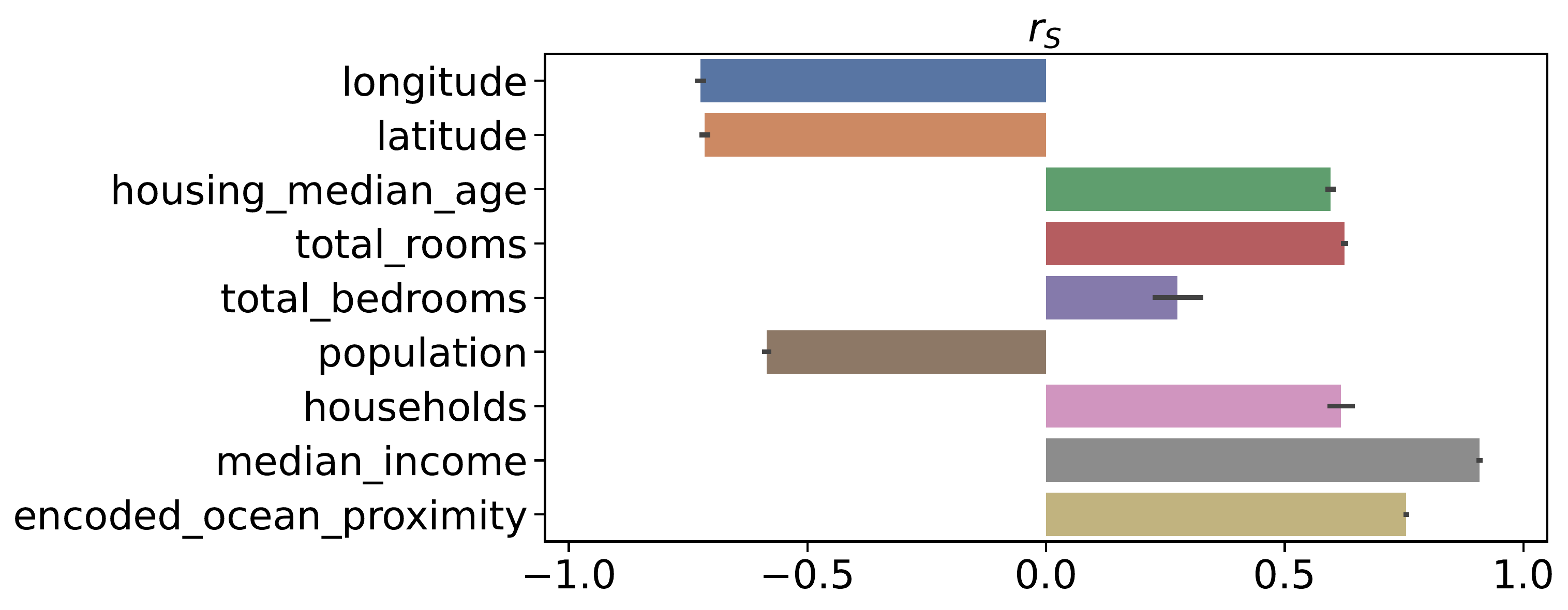} & \includegraphics[width=0.49\textwidth]{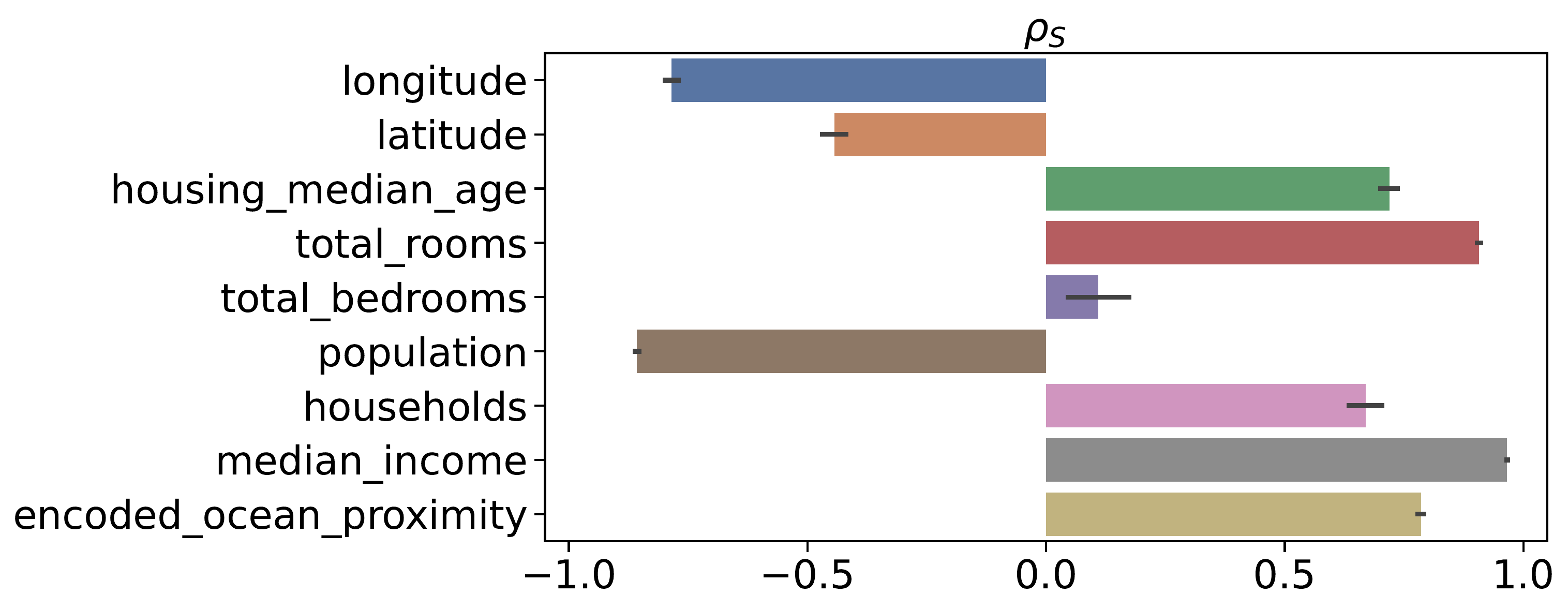}
        \tabularnewline
    \end{tabular}
    \caption{Comparison of the trend estimators on HOUSING. The linear model assigns a negative coefficient to the total number of rooms feature, even though the feature itself is positively correlated to the target.}\label{fig:barplot_housing}
\end{figure}

The random forests were run 100 times on HOUSING and the mean and standard deviation of the different trend estimators are shown in Fig. \ref{fig:barplot_housing}. 
The characteristic \emph{population} is very weakly negatively correlated with the housing price. 
However, all trend estimators report a significant negative trend for population. 
The feature total rooms is positively correlated with the target. 
However, the linear model assigns a negative coefficient to the total number of rooms. 
All other trend estimators report a positive trend. 

\subsection{Measures of Importance}
We compare the impurity-based feature importance, the permutation-based feature importance, and the feature importance induced by the two described residual algorithms (residual learning and Gram-Schmidt decorrelation).
A run consists of fitting a random forest.
To determine the residual-based importance scores, for each feature, 20 permutations that assign this feature to the last position are sampled independently.

\begin{figure}[ht]
    \centering
    \begin{subfigure}[b]{0.45\textwidth}
        \includegraphics[height=\textwidth]{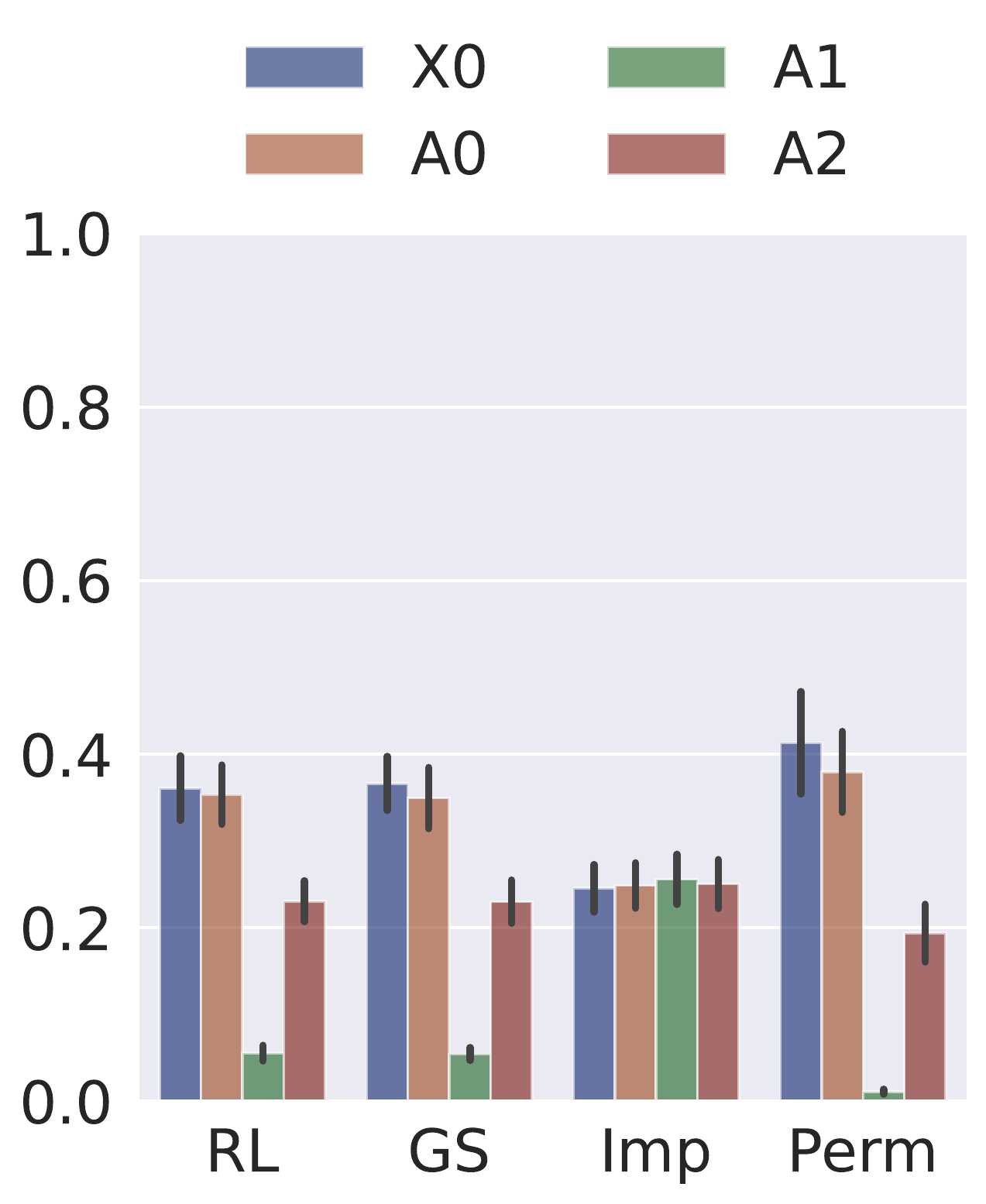}
    \end{subfigure}
    \begin{subfigure}[b]{0.45\textwidth}
        \includegraphics[height=\textwidth]{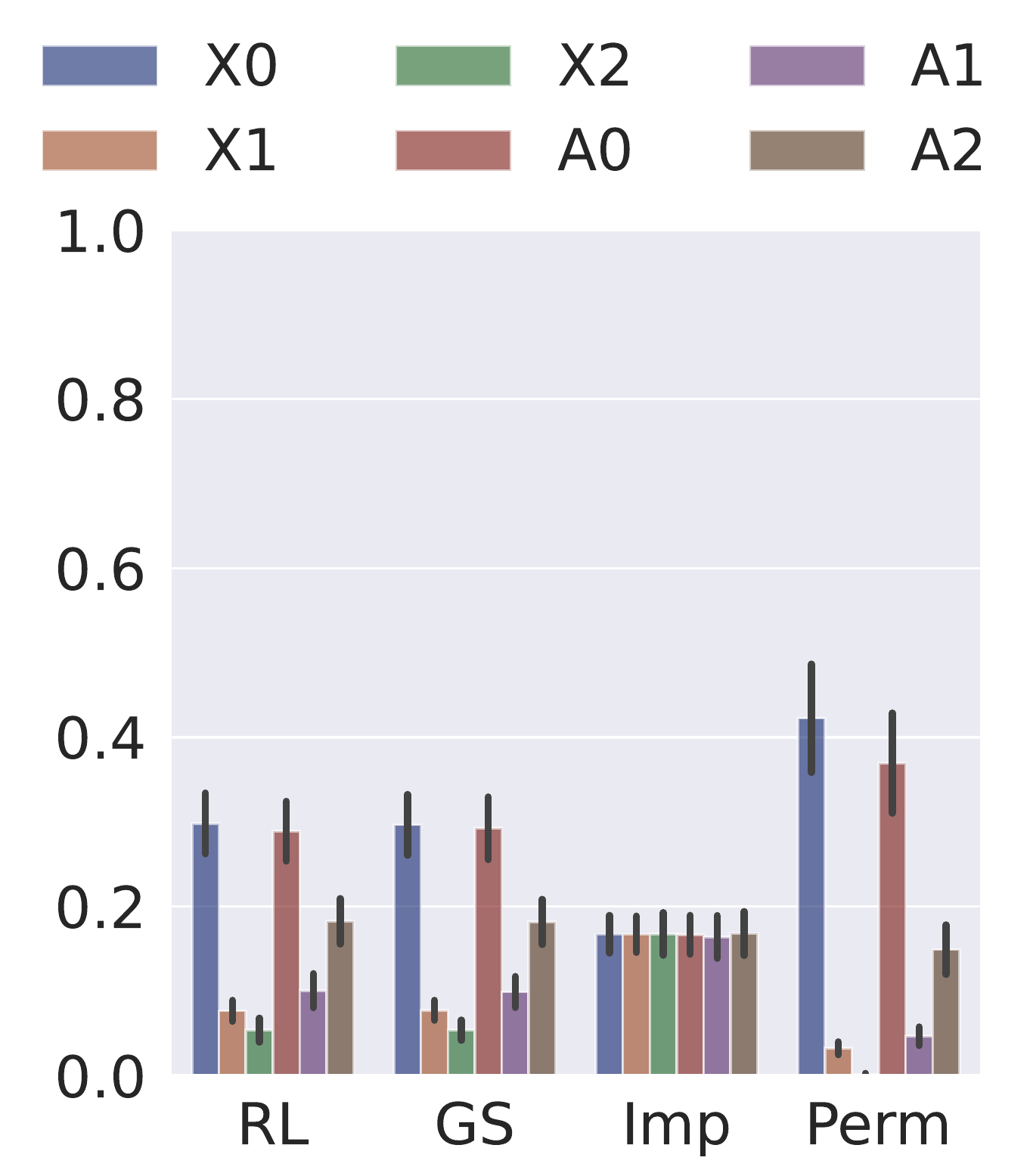}
    \end{subfigure} 
    \caption{Comparison of the four different notions of feature importance on synthetic data. The l.h.s. shows results with respect to SYN2. Here, the labels are generated as $Y = 4 \cdot X_0^{1.5}$, and $\{A_i\}$ are given as by $X_0 + \mathcal{W}_i$ for differently strong noise $\mathcal{W}_i$. On the r.h.s., results with respect to SYN3 are reported. Here, the labels are generated as $Y = 4 \cdot X_0^{1.5} + 2 \cdot X_1 + 0.5 \cdot X_2^2$, thus two more (weakly) informative features are given.}
    \label{fig:synImps}
\end{figure}

First, we compare the different scores on the synthetic datasets SYN2 and SYN3 (see Fig. \ref{fig:synImps}).
Perhaps the most important observation is that the impurity-based feature importance assigns the same score to all features -- in both datasets.
This is in strong contrast to all other feature importance scores.
It is noteworthy that both residual-based approaches produce very comparable scores on the given datasets.
Both residual-based approaches and the permutation-based score assign roughly the same score to $X_0$ and the slightly noisy variant $A_0$. 
However, feature $A_1$, which is subject to much more noise, receives a significantly higher score under residual-based scoring. 
Especially on SYN3, the residual-based approaches assign a not too small score to all informative features.
The permutation-based score for the informative features $X_1$ and $X_2$ is comparatively small.
However, all scores assign a higher importance to the noisy instance $A_0$ of $X_0$ than to the informative features $X_1$ and $X_2$.

\begin{figure}[ht]
    \centering
    \includegraphics[width=0.33\textwidth]{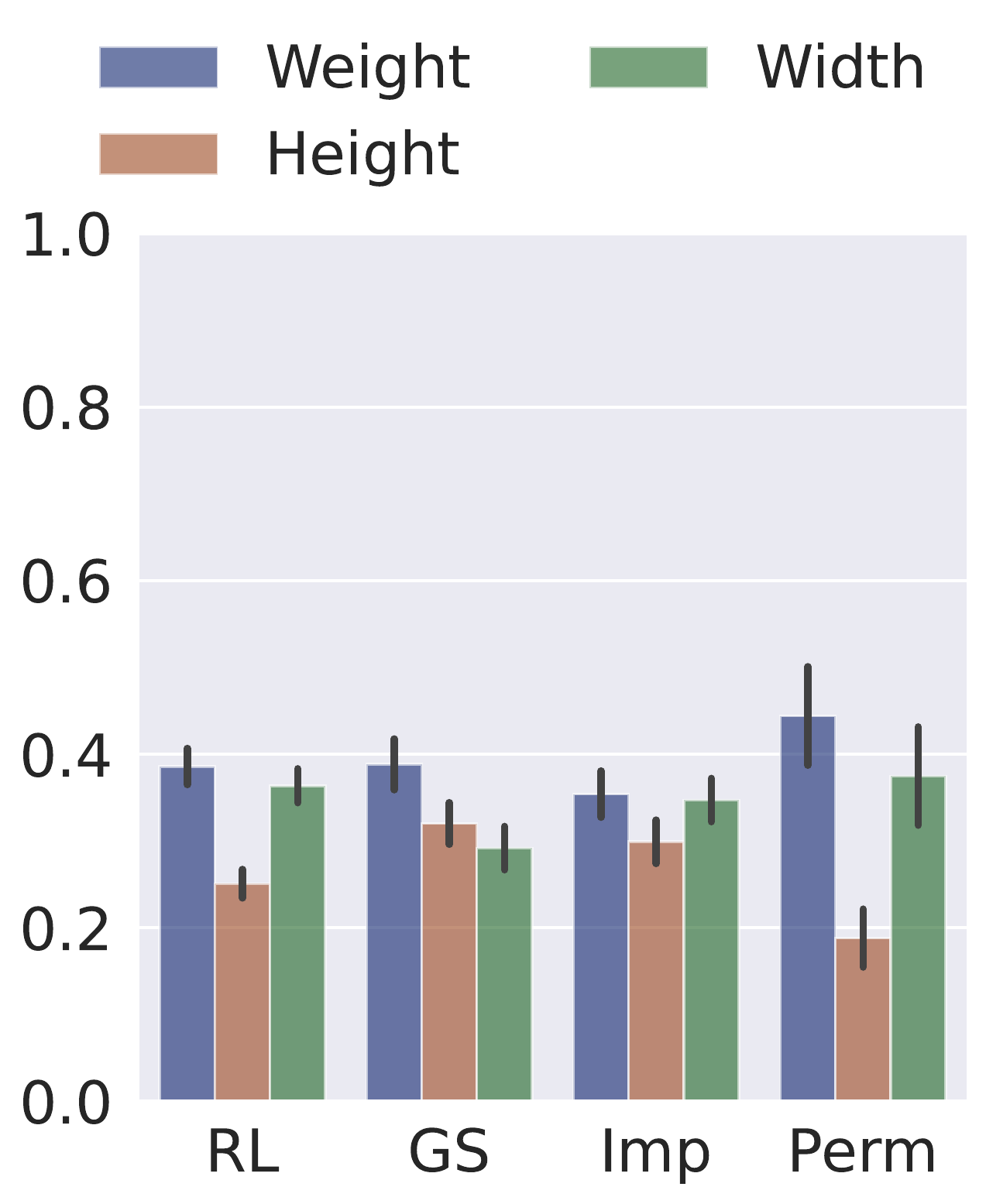}
    \includegraphics[width=0.657\textwidth]{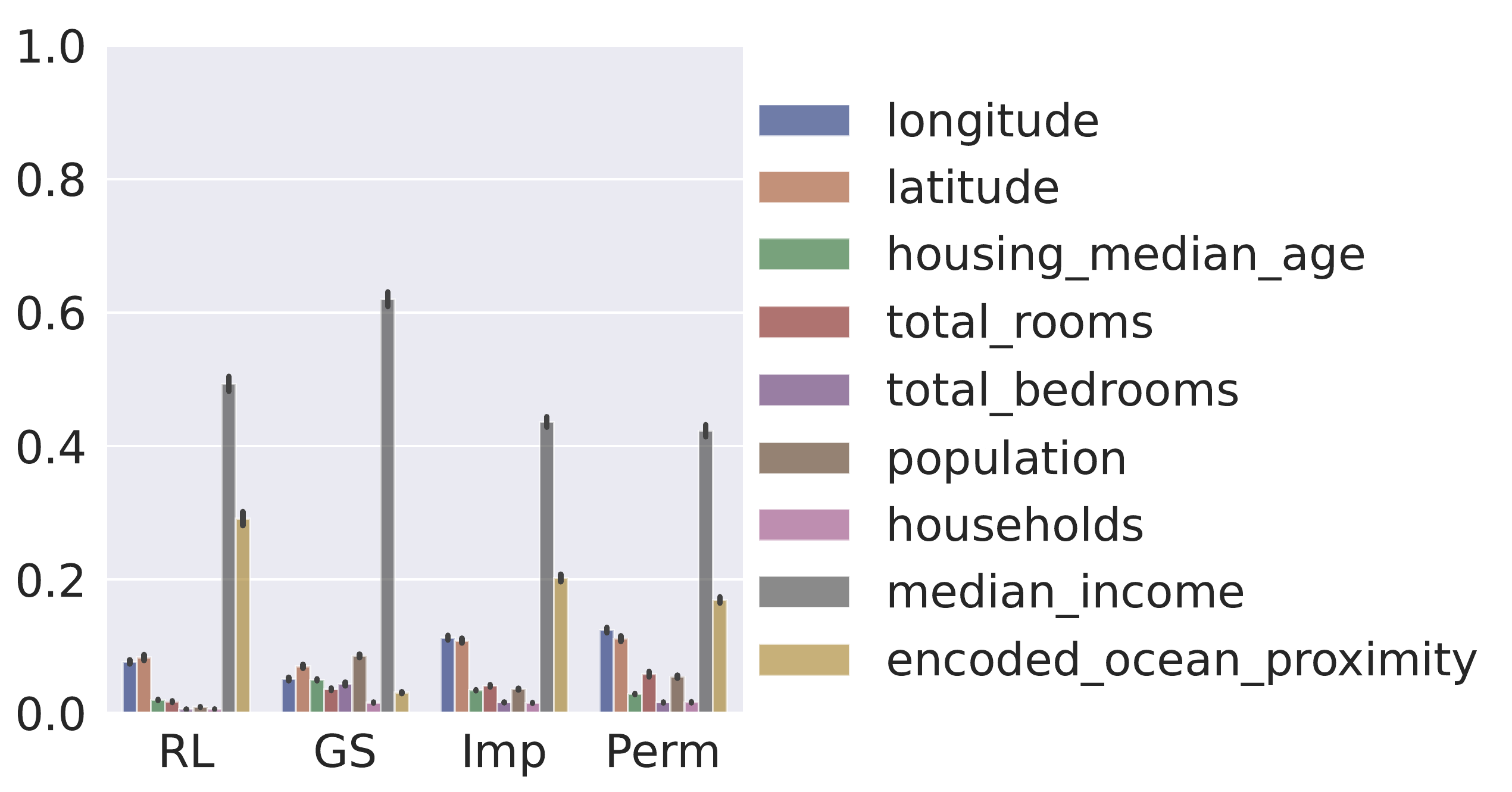}
    \caption{Comparison of the four different notions of feature importance on real-world instances. The l.h.s. reports the feature importance scores on the FISH dataset (mean and standard deviation over 400 independent runs), the r.h.s. on HOUSING (mean and standard deviation over 100 independent runs).}
    \label{fig:realdataImps}
\end{figure}

Next, we compare the different scores on the real data sets HOUSING and FISH (see Fig. \ref{fig:realdataImps}).
For HOUSING, it is most striking that the residual learning, impurity-based, and permutation-based scores assign the largest value to the median income, followed by the proximity to the ocean and the latitude/longitude, while the Gram-Schmidt-based score assigns only a large value to the median income and all other features receive comparable scores.
In addition, the population is found to be more important by the impurity-based and permutation-based approaches as opposed to the residual learning-based approach.

For FISH, all measures assign the highest score to weight and all measures assign a non-vanishing score to all three variables.
However, the score of Width is within one standard deviation in the residual learning-based, impurity-based, and permutation-based approaches.
Only the Gram-Schmidt-based score assigns a significantly larger value to Weight and considers Height to be the second most important feature.

\section{Conclusion}
We present two novel estimators for monotone trends in a dataset based on random forest regression.
They perform much more reliably than the often proposed linear model coefficient and are robust to noise.
However, the SHAP values perform equally well and are much better understood from a theoretical point of view.
Nevertheless, we believe that the transversal rate-based approach has its merits.
It depends only on the random forest model (trained on some dataset) and the computation is completely independent of the specific data, once that the model exists.
SHAP values, on the other hand, are computed as a combination of the model and some data (which may also have its own merits).

With respect to feature importance, we introduced the residual-based approach.
We compared the results on synthetic data and two real instances.
It is noteworthy that both residual-based approaches produce comparable results on the synthetic data sets, but this may be due to the fact that the noise is added linearly.
Overall, the residual-based approaches perform much better on highly correlated features than the impurity-based approach.
Their results are comparable to the permutation-based approach in many facets.
However, significant differences were also found.
In particular, informative features that contribute weakly to the noise were assigned higher values than by the permutation-based score.
Therefore, we believe that the residual-based feature importance scores should be preferred for use on datasets with highly dependent features.




\newpage
\bibliographystyle{plain}
\bibliography{bibliography}





\end{document}